%% file: main.tex
\documentclass[sigconf, nonacm]{acmart}
\AtBeginDocument{%
  }

\usepackage{bm}
\usepackage{amsthm}
\usepackage{algorithm}
\usepackage{algorithmic}
\theoremstyle{definition}

\newtheorem{definition}{\emph{Definition}}

\theoremstyle{remark}

\newtheorem{example}{Example}
\theoremstyle{definition}
\newtheorem{problem}{Problem}
\usepackage{graphicx}
\usepackage{subcaption}
\usepackage{pifont}
\usepackage{multirow}
\usepackage{multicol}

\newcommand{\model}{3dSAGER} 
\newcommand{\modelspace}{3dSAGER } 
\newcommand{\task}{geospatial ER} 
\newcommand{\taskspace}{geospatial ER } 
\newcommand{\taskspacecapital}{Geospatial ER } 
\newcommand{\fulltask}{geospatial ER for 3D objects} 
\newcommand{\fulltaskspace}{geospatial ER for 3D objects } 
\newcommand{\mesh}{mesh}
\newcommand{\meshspace}{mesh }

\newcommand*{\TechReport}{}

\newcommand{\revA}[1]{\textcolor{black}{#1}}
\newcommand{\revB}[1]{\textcolor{black}{#1}}
\newcommand{\revC}[1]{\textcolor{black}{#1}}
\newcommand{\revGeneral}[1]{\textcolor{black}{#1}}

\begin{document}


\title{3dSAGER: Geospatial Entity Resolution over 3D Objects (Technical Report)}

\author{Bar Genossar}
\affiliation{%
	\institution{Technion -- Israel Institute of Technology}
	\city{Haifa}
	\country{Israel}
}
\email{sbargen@campus.technion.ac.il}

\author{Sagi Dalyot}
\affiliation{%
	\institution{Technion -- Israel Institute of Technology}
		\streetaddress{P.O. Box 1212}
	\city{Haifa}
	\country{Israel}
		\postcode{43017-6221}
}
\email{dalyot@technion.ac.il}

\author{Roee Shraga}
\affiliation{%
	\institution{Worcester Polytechnic Institute}
	\city{Worcester, MA}
	\country{USA}
}
\email{rshraga@wpi.edu}

\author{Avigdor Gal}
\affiliation{%
	\institution{Technion -- Israel Institute of Technology}
	\city{Haifa}
	\country{Israel}
}
\email{avigal@technion.ac.il}

\begin{abstract}
Urban environments are continuously mapped and modeled by various data collection platforms, including satellites, unmanned aerial vehicles and street cameras. The growing availability of 3D geospatial data from multiple modalities has introduced new opportunities and challenges for integrating spatial knowledge at scale, particularly in high-impact domains such as urban planning and rapid disaster management.
Geospatial entity resolution is the task of identifying matching spatial objects across different datasets, often collected independently under varying conditions. Existing approaches typically rely on spatial proximity, textual metadata, or external identifiers to determine correspondence. While useful, these signals are often unavailable, unreliable, or misaligned, especially in cross-source scenarios.
To address these limitations, we shift the focus to the intrinsic geometry of 3D spatial objects and present \modelspace (3D Spatial-Aware Geospatial Entity Resolution), an end-to-end pipeline for geospatial entity resolution over 3D objects. \modelspace introduces a novel, spatial-reference-independent featurization mechanism that captures intricate geometric characteristics of matching pairs, enabling robust comparison even across datasets with incompatible coordinate systems where traditional spatial methods fail. As a key component of \model, we also propose a new lightweight and interpretable blocking method, BKAFI, that leverages a trained model to efficiently generate high-recall candidate sets.
We validate \modelspace through extensive experiments on real-world urban datasets, demonstrating significant gains in both accuracy and efficiency over strong baselines. Our empirical study further dissects the contributions of each component, providing insights into their impact and the overall design choices.
\end{abstract}


\maketitle
\input{introduction}
\input{problem_definition}

\input{model}

\input{hague_benchmark}

\input{experiments_new}

\input{related_work}

\input{conclusions}

\input{acknowledgements}


\newpage
\bibliographystyle{ACM-Reference-Format}
\bibliography{main}

\end{document}

%% file: introduction.tex
\begin{figure*}[t]
    \centering
    \includegraphics[width=.858\textwidth]{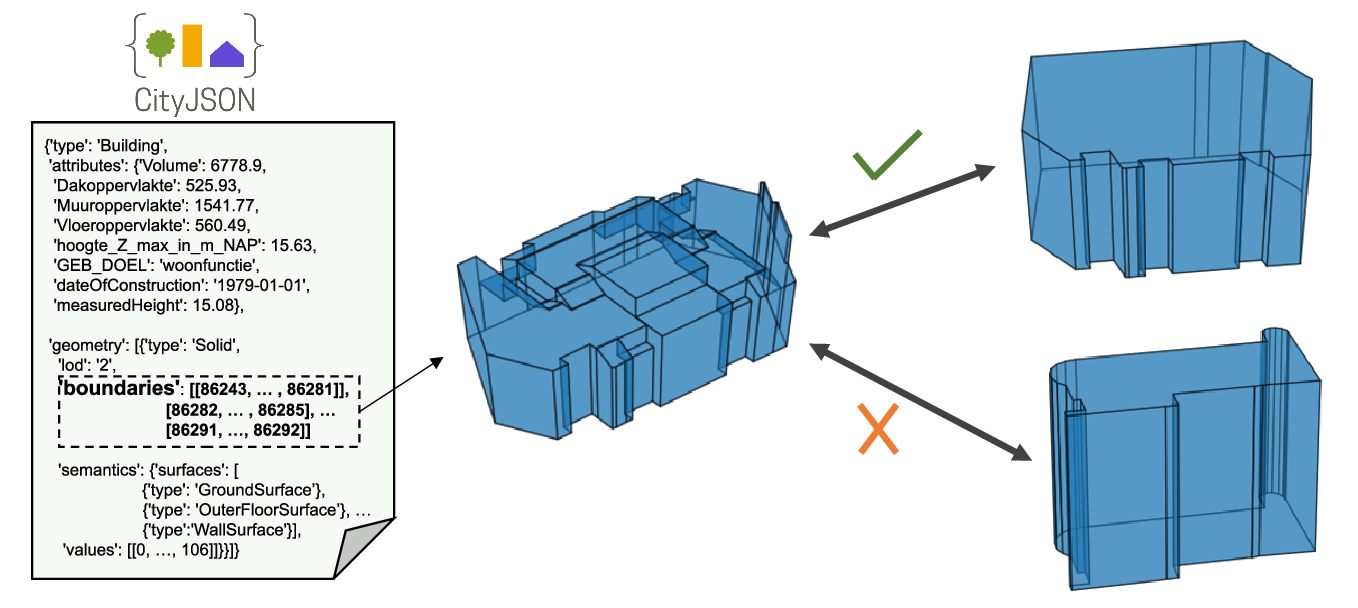}
    \caption{An example of \taskspace over 3D objects represented as CityJson files.}
    \label{fig:matching_example}
\end{figure*}

\section{Introduction}
\label{sec:introduction}
Entity Resolution (ER) is a data integration task, focusing on identifying and reconciling different data references that correspond to the same real-world entity~\cite{christen2012data, getoor2005link, elmagarmid2006duplicate, getoor2012entity, winkler2002methods}. \taskspacecapital has emerged as a unique branch of ER, focusing on the deduplication of geospatial objects~\cite{sehgal2006entity, balsebre2022geospatial, kang2007geoddupe, shah2021gem}. The task may involve different data modalities, such as textual descriptions and geographic coordinates.  
\begin{sloppypar}
\revB{In this work, we present \model, a novel end-to-end ER coordinate-agnostic pipeline that extends geospatial ER to the setting of 3D polygon mesh. \modelspace is designed for rich geometric representations such as 3D polygon meshes.} 
The work is motivated by the lack of reliable and accurate representations of urban environments. This is  particularly crucial in the context of urban disaster management\footnote{\url{https://www.wvi.org/disaster-management/urban-disaster-management}} that extensively relies on timely data~\cite{yu2018big}. 
Effectively collecting and processing such data involves understanding the heterogeneity of resources, automating the discovery and selection of data sources and dynamically reorganizing applications through model-driven feedback~\cite{balouek2022towards}.
Consider, for example, a system for early disaster warning, such as earthquakes and their subsequent events, {\em e.g.}, fires. Such systems need to identify rapidly evolving disaster situations where locality has a great impact. In this example, real-time fire spreading models used in Geographic Information System can identify the threatened areas and the physical structures, {\em e.g.} houses, that may be impacted.
Towards this goal, one can aim at integrating data from various sources, such as satellites, UAVs (Unmanned Aerial Vehicles), and street cameras to assist in coordinating emergency resources effectively~\cite{Adams2008TheAP}.
\end{sloppypar}

\begin{sloppypar}
Virtual 3D City Models (V3CMs) offer a detailed data representation technology, representing environmental and urban virtual information of surfaces and geospatial objects, including buildings, walls, windows, and vegetation~\cite{singh2013virtual,jovanovic2020building}.
Information on surfaces and objects is mostly stored in a standardized data model that allows format exchange of digital 3D models, among these are \emph{CityGML}~\cite{groger2012citygml} and \emph{CityJson}~\cite{ledoux2019cityjson}. The data model often combines textual descriptions ({\em e.g.}, location names) with coordinates ({\em e.g.}, latitude and longitude), and a textual representation of {\em polygon mesh}, a geometric object defined by its topology formulated according to a network of interconnected coordinates to form a closed 3D surface~\cite{bouzas2020structure, holzmann2017plane}. To provide reliable environmental data framework, the models stored in the V3CM are continuously updated as new data is acquired, which may lead to data duplications and conflicts when integrating or analyzing the information~\cite{erving2009data, dollner2007integrating}. 
Resolving these duplications is critical for ensuring an adequate and reliable representation of the environment.
\end{sloppypar}

Existing \taskspace methods consider coordinates in a pointwise fashion, where the geospatial description of an object is reduced to a single 
representative point, {\em e.g.}, its centroid~\cite{balsebre2022geospatial, shah2021gem, morana2014geobench, deng2019point}. Building on this, blocking, a preparatory step in ER that aims to discard non-matching pairs, often assumes that different references to the same entity should reside in nearby regions of the coordinate system. However, this assumption breaks when data sources employ different coordinate systems, leading to transformations that disrupt geometric alignment. Consequently, methods that rely on direct coordinate comparisons fail to generalize in such settings. 
Furthermore, other forms of geospatial information may be limited or unavailable, often due to urgent computing constraints that restrict the extraction of richer contextual data from images. Therefore, a coordinate-agnostic approach that leverages structural properties of objects becomes essential, \revB{motivating geometry-oriented featurization beyond 2D projections and metadata.}

\begin{example}
To illustrate the problem, consider  Figure~\ref{fig:matching_example}, which contains on the left an example of CityJson 3D object representation. On its right, a graphical illustration of the represented geospatial object (in this case, a building) is provided as polygon mesh. We also provide two additional object representations on the right.
 
Assume that we aim to match the object on the left to the two objects of the right. The top pair forms a match (marked with \ding{51}), due to their geometrical similarity (shape and boundary intricacies) of the two polygon meshes. In contrast, the bottom pair is identified as a non-match (marked as \ding{55}) due to irregularities, such as sharp edges and protrusions,  in their shapes.
\end{example}

Leveraging the complete polygon mesh for \fulltaskspace enables accurate matching based on intrinsic geometric and topological properties, rather than relying on pointwise representations accompanied with textual descriptions.
Therefore, we present \model, an end-to-end pipeline for \fulltaskspace that operates directly on polygon mesh representations of objects. We propose a novel featurization mechanism that transforms raw CityJson geometry into an interpretable feature space, capturing the intricate geometric characteristics between candidate matching pairs. \revB{Built upon a model trained over a subset of the data, we employ BKAFI, a lightweight and efficient coordinate-agnostic blocking technique that, to the best of our knowledge, is the first to leverage feature importance scores derived from a downstream matcher to select blocking keys. This approach departs from traditional approaches that extract blocking keys directly from the input data.}
The property vectors derived from our featurization process are subsequently used to construct pairwise feature vectors, which serve as input to a machine learning model, trained to determine whether a given candidate pair forms a match or not. Our approach is model-agnostic, allowing seamless integration with different models while ensuring both flexibility and interpretability. Our contribution can be summarized as follows.

\begin{itemize}
    \item A formal definition of \fulltaskspace and a corresponding end-to-end solution, \model, relying solely on the polygon mesh representations 
    without the use of any external metadata or location-based information.  
    \item A novel featurization framework for \task.
    \item A newly introduced blocking method, \emph{BKAFI}, which leverages feature importance to select a compact set of geometric properties -- offering a scalable and interpretable alternative to traditional coordinate-based blocking.
    \item \revA{A new large-scale benchmark dataset, \emph{The Hague}, built from real-world heterogeneous 3D sources.\footnote{\url{https://tinyurl.com/3dSAGERdataset}}}
    \item A comprehensive empirical evaluation demonstrating the effectiveness and efficiency of \modelspace. 
    \item Publicly available code to facilitate reproducibility.\footnote{\url{https://github.com/BarGenossar/3dSAGER}}
\end{itemize}

We provide basic notations and problem definition in Section~\ref{sec:problem}, followed by our solution proposal in Section~\ref{sec:model}. Section~\ref{sec:hague_benchmark} describes our new benchmark and Section~\ref{sec:experiments_full} outlines the empirical evaluation. Finally, related work is reviewed in Section~\ref{sec:related_work} and the paper is concluded in Section~\ref{sec:conclusions}.


%% file: problem_definition.tex
\section{Problem Definition}
\label{sec:problem}
We now define the task of \fulltask, \revB{performed over 3D polygon mesh data.}
Recall that CityJson is a common representation of such objects~\cite{ledoux2019cityjson} and thus, for the reminder of the paper we will assume that the 3D objects we have at our disposal are provided as CityJson files.
To establish a rigorous basis for our work, we begin by defining the concepts of a \textit{polygon} and a \textit{polygon mesh}, the fundamental representations of 3D objects in our setting. 

\begin{sloppypar}
\begin{definition}
\label{definition:polygon}
\emph{A {\em polygon} is a sequence of $n$ coordinates $p=\{v_1, v_2, \dots, v_n\}$, such that $v_i \in \mathbb{R}^3$ for every $v_i \in p$. 
The sequence forms a closed shape through a set of edges, expressed as an adjacency function  $\phi: \{v_1, \dots, v_n\} \times \{v_1, \dots, v_n\} \to \{0,1\}$, defined as:
\[
\phi(v_i, v_j) =
\begin{cases}
1, & \text{if } j = i+1 \text{ or } (i = n, j = 1)\\
0, & \text{otherwise}.
\end{cases}
\]
}
\end{definition}
\end{sloppypar}
The adjacency function ensures that edges exist between consecutive coordinates, including the edge connecting the last coordinate $v_n$ to the first vertex $v_1$, forming a closed surface.
With this definition, we next define a polygon mesh. 

\begin{definition}
\label{definition:polygon_mesh}
\emph{A {\em polygon mesh} (mesh for short) $P$ is a finite set of $m$ polygons (Definition~\ref{definition:polygon}) $P = \{ p_1, p_2, \dots, p_m \}$.
The polygons in $P$ collectively form a closed 3D sutrface, such that 
every edge $e$ belonging to a polygon $p_i \in P$ is shared by exactly one other polygon $p_j \in P$.}
\end{definition}

\begin{figure}[t]
    \centering
    \includegraphics[width=0.919\columnwidth]{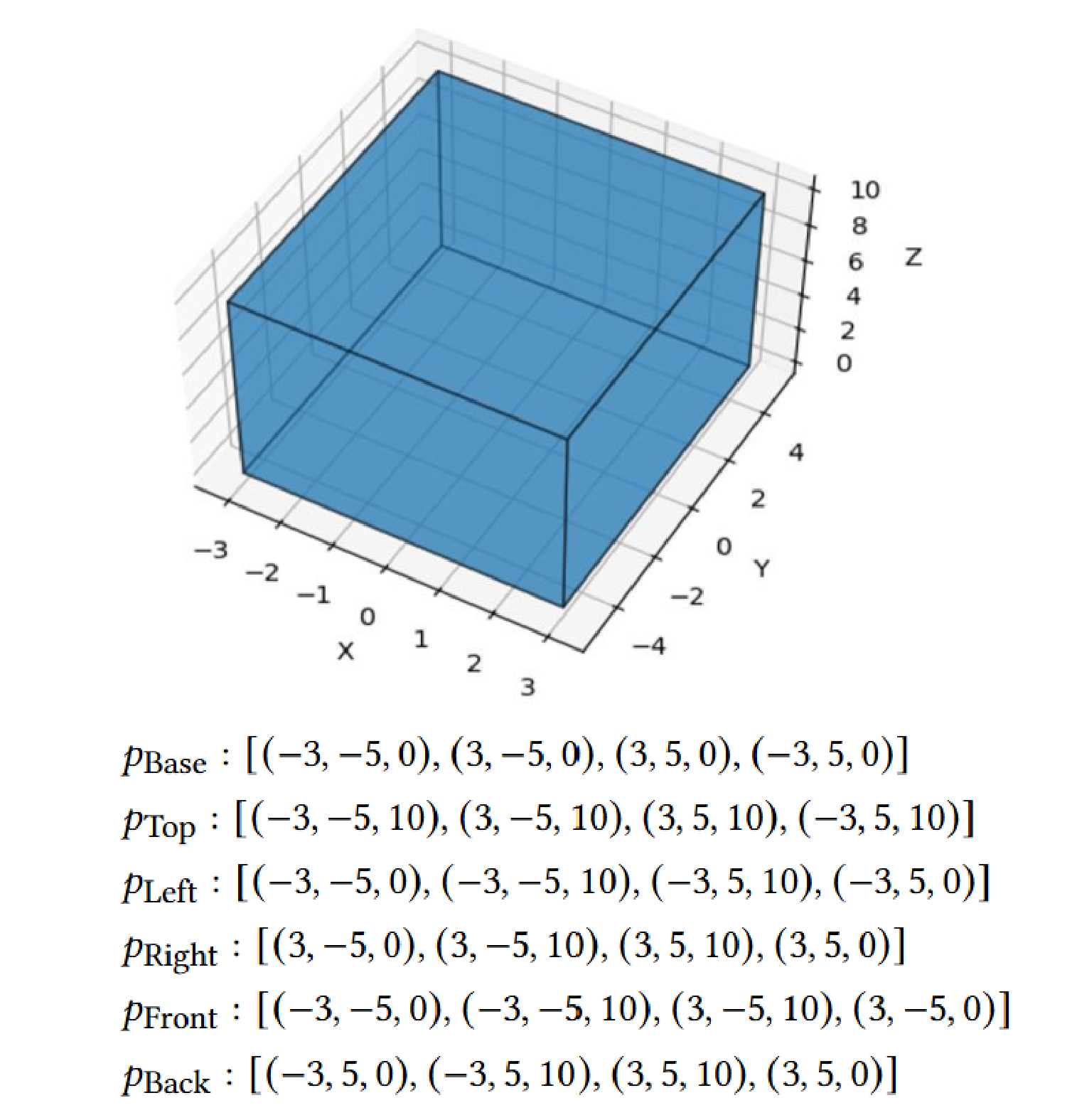}
    \caption{An example of a simple box-shaped \mesh, represented as a set of interconnected polygon faces.}
    \label{fig:polygon_mesh_example}
\end{figure}

\begin{example}
Figure~\ref{fig:polygon_mesh_example} provides an example of a simple box-shaped \meshspace (top) and the corresponding of six quadrilateral polygons (bottom), each defined by four coordinates in $\mathbb{R}^3$. 
The objects presented in  Figure~\ref{fig:matching_example} are also  \mesh es. The visualization was generated from  the `boundries' provided in the CityJson. 

\end{example}
\begin{sloppypar}
Equipped with the definition of a \mesh, we are now ready to define \task. 
Let $D^I=\{P^I_1,P^I_2,...,P^I_{n^I}\}$ and $D^C=\{P^C_1,P^C_2,...,P^C_{n^C}\}$ be sets of \mesh es, representing an indexed dataset and a candidate dataset, respectively.
We assume a clean-clean setting. This means that there are no matching objects internally in both $D^I$ and $D^C$, {\em i.e.}, $\nexists P^I_i,P^I_j\in D^I$ ($\nexists P^C_i,P^C_j\in D^C$) such that $P^I_i$ ($P^C_i$) and $P^I_j$ ($P^C_j$) refer to the same real-world entity. 
The ER objective is to identity objects, in this case 3D objects represented as \mesh es, that refer to the same \emph{real-world entity}.  In the context of our work, such entities can be buildings or other geospatial objects that do not have unique identifiers. Following the common literature for ER~\cite{christen2012data}, we address \taskspace as a two stage problem consisting of \emph{blocking} and \emph{matching}. Largely speaking, the former aims to reduce the number of pairwise object comparisons, since comparing all possible $|D^I|\times|D^C|$ pairs can be prohibitively expensive. An effective blocking strategy balances the goal of retaining as many true matches as possible while minimizing the number of non-matching pairs. The matching stage is typically more resource-intensive, as it incorporates nuanced characteristics for better differentiation between matching and non-matching pairs. Stemming mainly from the absence of primary keys, matching relies on semantic cues, may they be geospatial or geometric, to asses whether polygons refer to the same real-world spatial object. 
\end{sloppypar}

\noindent\textbf{Geospatial ER Blocking:}
Given an indexed dataset $D^I$ and a candidate dataset $D^C$, blocking step $B(D^I, D^C)$ (or a \emph{blocker}) aims to discard obvious non-matching \meshspace pairs, thereby reducing the computational complexity of subsequent steps. The outcome of this step is a set of candidate \meshspace pairs $C_{ D^I\times D^C} \in D^I\times D^C$ ($C$ when clear from context).

\begin{figure*}[t]
    \centering
    \scalebox{0.81}{\includegraphics[width=\textwidth]{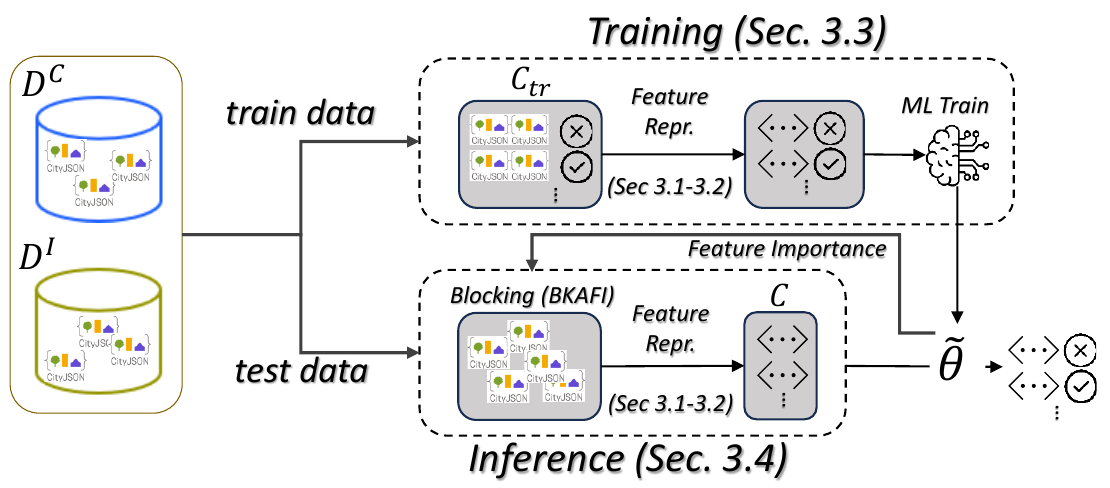}}
    \caption{A schematic overview of the full pipeline.}
    \label{fig:pipeline}
\end{figure*}

\noindent\textbf{Geospatial ER Matching:}
Given a blocker $B(D^I, D^C)$, we now define the matching problem. We assume the existence of an oracle mapping function $\theta$, unknown to us. That is, our goal is to obtain a model $\tilde{\theta}$ to identify duplicate (matching) \mesh es in $C$.

\begin{problem}
Let $D^I=\{P^I_1,P^I_2,...,P^I_{n^I}\}$ be an index dataset, $D^C=\{P^C_1,P^C_2,...,P^C_{n^C}\}$ a candidate dataset and $C \in D^I\times D^C$ a set of \meshspace pairs obtained by a blocker $B(D^I, D^C)$. The matching problem aims to obtain a model $\tilde{\theta}: C\rightarrow \{0,1\}$, classifying a pair $(P^I_i, P^C_j)\in C$ as a match ($1$) or non-match ($0$).
\end{problem}

The focus of our work is in developing (1) a blocker $B$ and (2) a model $\tilde{\theta}$. Specifically, the blocker is based on a reduced set of features that support maximum recall at a low cost. The model is a machine-learning model, trained on pairwise feature vectors that encode the geometric relationships between \mesh es in candidate pairs. Obviously, a ``perfect'' $\tilde{\theta}$ is one that is equivalent to $\theta$ (or at least has equivalent outputs). For the context of this work we assume we have at our disposal a set that was annotated with $\theta$ (aka a training set), which can be used to train a model. For simplicity of presentation, this study focuses on clean-clean ER assumption. 
Yet, the proposed framework is flexible and can be adapted to non clean-clean settings as well. These include settings where duplicates may exist within the indexed dataset $D^I$, the candidate dataset $D^C$, or both, enabling broader applicability in diverse real-world contexts.

\revC{To illustrate the relevance of this setting, we introduce next a case study of a post-disaster management (PDM) scenario. Natural disasters ({\em e.g.}, earthquakes and floods) require rapid and accurate situational awareness for prioritizing emergency response. A central challenge in such scenarios, especially in urban areas, involves aligning newly captured 3D imagery of damaged areas, often collected by UAVs, against a registry of existing 3D images, as maintained by local authorities.
The UAV-acquired datasets, represented by \mesh es, may lack trustworthy geo-location due to factors such as GPS disruption or the lack of correction infrastructure (e.g., CORS networks)~\cite{zhuo2017automatic,manzini2024crasar}, particularly in remote or underdeveloped regions~\cite{verykokou2016uav,chiabrando2019uav,xu2021volumetric}. Moreover, imagery metadata may be missing or inconsistent~\cite{ulusoy2014image}, making traditional matching signals like spatial proximity or external identifiers ineffective. The only reliable information may be the intrinsic geometric structure of captured objects. As a part of our empirical analysis (Section~\ref{subsection:mra}), we analyze the role of matching in supporting PDM.}

%% file: model.tex
\section{\model: Matching 3D Objects}
\label{sec:model}

We now present \modelspace (3D Spatial-Aware Geospatial Entity Resolution), our novel end-to-end solution for geospatial ER over 3D objects. As depicted in Figure~\ref{fig:pipeline}, \modelspace comprises two main stages, namely \emph{training} (top panel) and \emph{inference} (bottom panel). Both stages operate over the indexed dataset $D^I$ and the candidate dataset $D^C$. In the initial step, \meshspace items from both datasets are transformed into \emph{property vectors} (Section~\ref{subsubsection:property_vectors}), \revB{which are extracted directly from the polygon mesh geometry, unlike traditional ER pipelines that rely on tabular or metadata-derived features. This featurization is structurally tied to the 3D mesh representation and designed to capture intrinsic geometric characteristics relevant for matching. To the best of our knowledge, this is the first application of such a featurization approach in the context of ER.}
In the training stage, we use a set of labeled object pairs $C_{tr} \subseteq D^I \times D^C$, represented as property vectors. These pairs undergo a featurization step that converts them into a machine-learning-compatible format, capturing pairwise geometric relationships between the objects (Section~\ref{subsection:feature_vectors}). This representation is then leveraged to train an interpretable binary classifier $\tilde{\theta}$, such as XGBoost~\cite{chen2016xgboost}\footnote{Gradient-boosted decision trees, {\em e.g.}, XGBoost, are considered state-of-the-art for tabular data~\cite{mcelfresh2023neural}.} (Section~\ref{subsection:train}). The output of this phase is the trained model $\tilde{\theta}$.
The inference phase resembles a traditional ER pipeline: a blocking method (in our case, the proposed \emph{BKAFI}, Section~\ref{subsection:blocking}) is first applied to the input datasets, generating a candidate set $C \subseteq D^I \times D^C \setminus C_{tr}$ (excluding training pairs). This filtered set is then passed to the matching phase for final prediction (Section~\ref{subsection:matching}).

\subsection{Property Vectors}
\label{subsubsection:property_vectors}
We introduce first the generation of property vectors. Raw \mesh es, typically expressed as sequences of 3D coordinates, are not inherently interpretable by machine learning models. Even for human experts, analyzing raw coordinate data provides limited insight into the structure or semantics of spatial objects. This low-level representation lacks the abstraction needed to capture meaningful geometric relationships.
While visualization tools (see, {\em e.g.}, Figures~\ref{fig:matching_example} and~\ref{fig:polygon_mesh_example}) enable humans to perceive subtle characteristics and nuances within polygons, translating these insights into machine-readable features requires a systematic computational approach. To address this challenge we propose to generate a property vector that encodes various geometric properties of a \mesh, enabling its processing by machine learning algorithms later in the pipeline.

\begin{sloppypar}
The full set of computed geometric properties, denoted by $G = \{g_1, g_2, \dots, g_n\}$, captures a diverse range of characteristics describing each \meshspace from multiple perspectives~\cite{labetski20233d}. These properties encompass dimensions, shape descriptors, boundary complexity, and structural indicators -- together providing a comprehensive geometric profile.
Table~\ref{table:properties} presents a selected subset of representative properties used in our pipeline. The rightmost columns of the table provide examples of a candidate mesh from $D^C$ (left) and an object from the index set $D^I$ (right). 
\ifdefined\TechReport

\else
The complete list of properties, along with detailed descriptions are available in a technical report.
\fi
\end{sloppypar}

For each \meshspace $p$ in $D^I$ and $D^C$, we compute a property vector $G_p = (g_1(p), g_2(p), \dots, g_n(p))$, where each $g_i(p)$ corresponds to a specific geometric property. The columns \texttt{Cand. Example} and \texttt{Ind. Example} in Table~\ref{table:properties} provide the actual computed values for the real matching pair illustrated in Figure~\ref{fig:matching_example}, where the middle object originates from the candidate dataset $D^C$, while the object on the right comes from the indexed dataset $D^I$. Although these two objects differ visually and numerically in several geometric properties, they are still correctly identified as a match. For instance, the \texttt{num\_vertices} property varies significantly (194 vs. 48), while others such as \texttt{perimeter} and \texttt{convex\_hull\_area} remain closely aligned. These variations illustrate the role of our property-based approach in tolerating moderate differences while leveraging consistent structural signals across sources.
To mitigate the effect of scale disparities across properties while preserving relative magnitudes, we apply a logarithmic normalization to the computed features using the $\log(1 + x)$ transformation~\cite{changyong2014log, manning2001estimating}. Formally, we normalize a property value $g_i(p)$ for a mesh $p$ to be $\tilde{g}_i(p) = \log(1 + g_i(p))$. 

\ifdefined\TechReport
\begin{table}[t]
\centering
\caption{Object Properties Computed over each Polygon. The \texttt{Ind. Example} and \texttt{Cand. Example} columns contain the  property values computed for the matching objects in Figure~\ref{fig:matching_example}.}
\scalebox{0.5825}{\begin{tabular}{|c|c|c|c|}
\hline
\multicolumn{2}{|l|}{\textbf{1. Size and Dimensions}} & \textbf{Cand. Example} & \textbf{Ind. Example} \\
\hline
\texttt{area} & Total surface area of the mesh. & 292412 & 271745 \\
\texttt{volume} & 3D volume of the mesh. & 6778859 & 1269.7 \\
\texttt{height\_diff} & Height difference between vertices. & 29.7 & 11.6 \\
\texttt{num\_vertices} & Number of unique vertices & 194 & 48 \\
 & defining the mesh. &  &  \\
\hline
\multicolumn{2}{|l|}{\textbf{2. Perimeter and Boundary Characteristics}} &  &  \\
\hline
\texttt{perimeter} & Total length of polygon's boundary. & 114043.9 & 114003.7 \\
\texttt{circumference} & Boundary length. & 8.6 & 5.2 \\
\texttt{perimeter\_index} & Ratio of perimeter to area & 1.7 & 1.6 \\
 & indicating boundary complexity &  &  \\
\hline
\multicolumn{2}{|l|}{\textbf{3. Shape Complexity and Geometry}} &  &  \\
\hline
\texttt{convex\_hull\_area} & Area of the polygon's convex hull & 108013.9 & 107877.8 \\
 & comparing irregularity. &  &  \\
\texttt{ave\_centroid\_distance} & Average distance of boundary & 14077.1 & 16336.8 \\
 & points from the centroid. &  &  \\
\texttt{shape\_index} & Ratio reflecting compactness of the shape. & 4.1 & 5.6 \\
\texttt{fractality} & Self-similarity, indicating shape irregularity.
& 0.0 & 0.0 \\
\texttt{elongation} & Degree of shape's elongation. & 2.8 & 3.1 \\
\texttt{hemisphericality} & Degree of similarity to a hemisphere. & 17.8 & 8.3 \\
\texttt{cubeness} & Degree of similarity to a cube. & 10.6 & 6.4 \\
\hline
\multicolumn{2}{|l|}{\textbf{4. Symmetry and Alignment}} &  &  \\
\hline
\texttt{axes\_symmetry} & Symmetry relative to principal axes. & 6967.7 & 8854.1 \\
\hline
\multicolumn{2}{|l|}{\textbf{5. Structural and Functional Properties}} &  &  \\
\hline
\texttt{density} & Compactness or mass distribution & 25640.3 & 23836.6 \\
 & within the polygon's area/volume. &  &  \\
\texttt{num\_floors} & Number of distinct floors. & 25 & 49 \\
\hline
\end{tabular}
}
\label{table:properties}
\end{table}

\else
\begin{table}[t]
\centering
\caption{Object Properties Computed over each Polygon. The \texttt{Ind. Example} and \texttt{Cand. Example} columns contain the  property values computed for the matching objects in Figure~\ref{fig:matching_example}.}
\scalebox{0.625}{\begin{tabular}{|c|c|c|c|}
\hline
\textbf{Property} & \textbf{Description} & \textbf{Cand. Example} & \textbf{Ind. Example} \\
\hline
\texttt{area} & Total surface area of the mesh. & 2924125119 & 2717459053 \\
\texttt{num\_vertices} & Number of unique vertices & 194 & 48 \\
 & defining the mesh. &  &  \\
\texttt{perimeter} & Total length of polygon's boundary. & 114043.9 & 114003.7 \\
\texttt{perimeter\_index} & Ratio of perimeter to area & 1.7 & 1.6 \\
 & indicating boundary complexity &  &  \\
\texttt{convex\_hull\_area} & Area of the polygon's convex hull & 108013.9 & 107877.8 \\
 & comparing irregularity. &  &  \\
\texttt{ave\_centroid\_distance} & Average distance of boundary & 14077.1 & 16336.8 \\
 & points from the centroid. &  &  \\
\texttt{elongation} & Degree of shape's elongation. & 2.8 & 3.1 \\
\texttt{axes\_symmetry} & Symmetry relative to principal axes. & 6967.7 & 8854.1 \\
\texttt{density} & Compactness or mass distribution & 25640.3 & 23836.6 \\
 & within the polygon's area/volume. &  &  \\
\hline
\end{tabular}
}
\label{table:properties}
\end{table}

\fi

\subsection{Pairwise Feature Vectors}
\label{subsection:feature_vectors}
Our goal is to identify matches between objects despite property discrepancies, leveraging both shared and differing geometric characteristics.
With a property vector representing a single \mesh, 
the next step is to generate feature vectors that encapsulate the pairwise relationships between corresponding object properties. Formally, we use $F(P_1, P_2)$ to denote a feature vector over the pair of \mesh es $P_1 \in D^I$ and $P_2 \in D^C$.
To capture the mutuality of object properties, we introduce a \emph{division} operator that computes the element-wise ratio between corresponding entries in the two property vectors.

\begin{definition}[Division Operator]
\label{definition:div}
\begin{sloppypar}
For \mesh es $P_1 \in D^I$ and $P_2 \in D^C$, let $G_{P_1} = (g_1(P_1),g_2(P_1),\dots,g_n(P_1))$ be a property vector of $P_1$ and  
$G_{P_2} = (g_1(P_2), g_2(P_2),\dots,g_n(P_2))$ be a property vector of $P_2$.  
The {\em division operator}, denoted by $G_{P_1} \div G_{P_2}$, is defined as follows.
\end{sloppypar}
\begin{equation}
    F_{\div}(P_1, P_2) = G_{P_1} \div G_{P_2}
\end{equation}
where
\begin{equation}\nonumber
    F_{\div}(P_1, P_2)[i] = \frac{g_i(P_1)}{g_i(P_2)}, \quad \forall g_i(P_1) \in G_{P_1}, \quad g_i(P_2) \in G_{P_2}.
\end{equation}
\end{definition}

An entry in the feature vector $F_{\div}(P_1, P_2)$ encodes the proportional variation of a property in the two objects. By expressing pairwise relationships as ratios (rather than absolute differences), this representation remains meaningful even when the objects originate from sources with differing scales or systematic offsets, thereby enhancing robustness to discrepancies across data sources.

\vspace{.1in}\noindent\textbf{Systematic Discrepancy:} Intuitively, one might expect that matching \meshspace references should exhibit closely aligned property values. In such cases, the ratios computed by $F_{\div}(\cdot)$ would be approximately 1. However, since the datasets originate from distinct sources with varying acquisition methods, timestamps, and modeling conventions, consistent patterns of variation may arise, referred to as systematic discrepancy. 
As all \mesh es in $D^C$ come from the same source, they may collectively deviate from those in $D^I$ in specific properties. For instance, one dataset may systematically underestimate dimensions such as area or height, or apply different resolutions or coordinate transformations. These deviations are not random noise, but rather reflect structured biases introduced during data collection. We formally define systematic discrepancy next.

\begin{definition}[\emph{$(\epsilon, \delta)$-Systematic Discrepancy}]\label{definition:systematic_discrepancy}
Let $D^I$ and $D^C$ denote the indexed and candidate sets, respectively, and let $g \in G$ be a geometric property. We say that $D^I$ and $D^C$ exhibit an \emph{$(\epsilon, \delta)$-systematic discrepancy} with respect to $g$ if there exists a ratio $r_g \in \mathbb{R}^+$ such that at least a $(1 - \delta)$ fraction of the known matching pairs $(P_1, P_2)$, with $P_1 \in D^I$ and $P_2 \in D^C$, satisfy
\begin{equation}
r_g - \epsilon \leq \frac{g(P_1)}{g(P_2)} \leq r_g + \epsilon.
\end{equation}
\end{definition}

$\epsilon$ represents the tolerance allowed for the ratio of geometric property values between matching pairs. A smaller value of $\epsilon$ indicates a stricter condition for the matching pairs, implying that the properties of the pairs should be more similar to be considered a valid match. 
$\delta$ reflects the efficacy of the tolerance condition. 
A smaller value of $\delta$ (closer to 0) implies that a higher proportion of matching pairs must meet the condition, resulting in a stricter requirement for the consistency of the geometric properties.

Another perspective on $(\epsilon, \delta)$-systematic discrepancy is through the lens of statistical dispersion. Given a set of ratios $\left\{ \frac{g(P_1)}{g(P_2)} \right\}$ for matching pairs $(P_1, P_2)$ and a property $g$, the standard deviation $\sigma$ of this ratio distribution provides an empirical way to estimate a reasonable $\epsilon$ value. In particular, if the ratios are normally distributed, setting $\epsilon = 2\sigma$ would cover roughly 95\% of the values, corresponding to $\delta \approx 0.05$. Thus, the standard deviation offers a data-driven approach to quantify and validate systematic discrepancy, with lower $\sigma$ implying stronger consistency across sources. 

The property values in Table~\ref{table:properties} exemplify the phenomenon of \emph{systematic discrepancy}. For instance, the \texttt{convex\_hull\_area} property only slightly differs between the candidate (108{,}013.9) and indexed (107{,}877.8) objects. Conversely, properties like \texttt{axes\_symmetry} exhibit more noticeable differences, despite the objects being a confirmed match (Figure~\ref{fig:matching_example}). Such patterns, when consistently observed across numerous matching pairs, indicate systematic (rather than random) discrepancies.
While certain ratios may be inferred through direct computation over the training data, more intricate relationships between different properties are not always apparent through simple observation. Instead, we delegate the identification of such geometric consistencies to a learning algorithm, which can effectively capture and leverage these complex patterns.

\subsection{\modelspace Training Phase}
\label{subsection:train}
The feature vectors of objects in the training set $C_{tr}$ serve as input to a machine learning model $\tilde{\theta}$ for classification. During training, the model learns to map the input feature vectors to binary labels, where a label of $1$ indicates that the \meshspace pair corresponds to the same real-world entity, and $0$ otherwise. The training process optimizes the model parameters by minimizing a loss function that measures the variation between the predicted and true labels.
\modelspace is machine learning model-agnostic, offering a flexible and adaptable framework that can be incorporated into any model. We emphasize the use of interpretable models, enabling the utilization of feature importance scores in the blocking phase.

\subsection{\modelspace Inference Phase}
Equipped with a trained model $\tilde{\theta}$, we now describe the inference process for blocking (Section~\ref{subsection:blocking}) and matching (Section~\ref{subsection:matching}). 

\subsubsection{Blocking based on Feature Importance}
\label{subsection:blocking}

\begin{figure}[t]
    \centering
    \scalebox{0.995}{\includegraphics[width=\columnwidth]{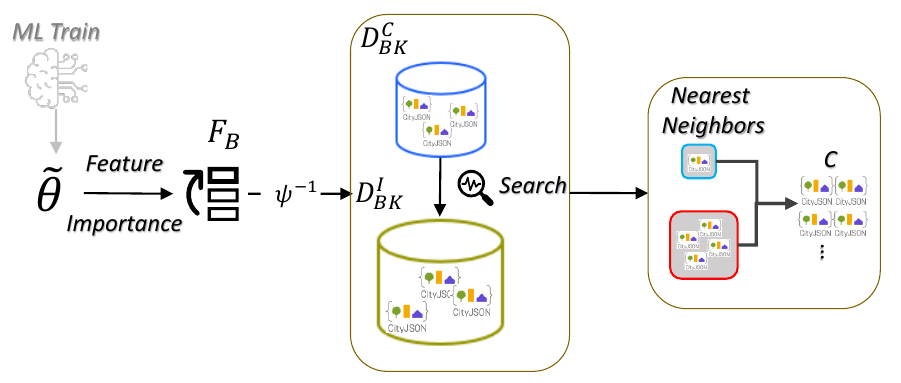}}
    \caption{An illustration of BKAFI.} 
    \label{fig:pipeline_blocking}
\end{figure}

We introduce a new blocking method, \emph{Blocking Key as Feature Importance} (\emph{BKAFI}), tailored for \fulltask. The key insight is that a small subset of geometric properties captures most of the information needed for matching. Hence, instead of searching the full feature space, BKAFI restricts the search to a space defined by the most informative features. To identify these, we trace back the top-ranked features from the trained model $\tilde{\theta}$ to their originating geometric properties (Section~\ref{subsection:feature_vectors}). For each candidate object, we then retrieve similar objects from $D^I$ based on their values in this reduced space.
\revB{BKAFI introduces several key differences from traditional blocking methods. It avoids reliance on coordinate data, operates over features extracted from polygon mesh geometry, and, critically, derives blocking keys from the trained matcher’s feature importance scores rather than manually selected attributes. This approach not only improves adaptability to non-standard 3D \meshspace data, but also ensures that the blocking process remains lightweight and efficient.}

\begin{algorithm}[t]
\caption{BKAFI: Blocking Key as Feature Importance.}
\label{alg:bkafi}
\begin{algorithmic}[1]
\REQUIRE Trained model $\tilde{\theta}$, candidate set $D^C$, $|F_{B}|$ search space size, index set $D^I$, nearest neighbor parameter $k$

\STATE Initialize modified candidate set $D^C_{BK} \gets \emptyset$ \label{line:init_D_C_BKAFI}
\STATE Initialize modified indexed set $D^I_{BK} \gets \emptyset$ \label{line:init_D_I_BKAFI}
\STATE Initialize set of candidate \mesh pairs $C \gets \emptyset$ \label{line:init_C}
\STATE $F_{B} \gets$ Extract top $|F_{B}|$ features with the highest feature importance scores from $\tilde{\theta}$ \label{line:extract_features}
\STATE $G^{\prime} \gets \{\psi^{-1}(f) \forall f \in F_{B} \}$ \label{line:map_features}

\FOR{each $P \in D^C$}
    \STATE $G_{P_{BK}} \gets G_{P}[G^{\prime}]$ \label{line:extract_P}
    \STATE Add $G_{P_{BK}}$ to $D^C_{BK}$ \label{line:add_to_D_C}
\ENDFOR
\FOR{each $P \in D^I$}
    \STATE $G_{P_{BK}} \gets G_{P}[G^{\prime}]$ \label{line:extract_P_index}
    \STATE Add $G_{P_{BK}}$ to ~$D^I_{BK}$ \label{line:add_to_D_I}
\ENDFOR

\STATE $BKAFI_{Ind} \gets Index (D^I_{BK})$ \label{line:index_D_I}
\FOR{each $G_{P_{BK}} \in D^C_{BK}$}
    \STATE $P \gets $ Map $G_{P_{BK}}$ to its corresponding item in $D^C$ \label{line:map_back_P}
    \STATE $k_{G_{P_{BK}}} \gets \text{Retrieve}(k, G_{P_{BK}}, BKAFI_{Ind})$ \label{line:retrieve_neighbors}
    \FOR{each $G_{P_{I}} \in k_{G_{P_{BK}}}$}
        \STATE Map $G_{P_{I}}$ back to its corresponding item in $P_I \in D^I$ and add $(P, P_I)$ to $C$ \label{line:add_to_C}
    \ENDFOR
\ENDFOR

\RETURN $C$ \label{line:return_C}
\end{algorithmic}
\end{algorithm}

The \emph{BKAFI} algorithm (see Algorithm~\ref{alg:bkafi} for pseudocode and Figure~\ref{fig:pipeline_blocking} for illustration) receives as input a trained model $\tilde{\theta}$, candidate set $D^C$, and index set $D^I$. $F_{B}$ denotes the subset of features of size ($|F_{B}|$). $k$ is the number of candidate pairs generated per each object in $D^C$, which is also provided to the algorithm as input. 

The algorithm extracts the $|F_B|$ features with the highest feature importance scores from $\tilde{\theta}$ (Line~\ref{line:extract_features}). Next, the original geometric properties corresponding to the features in $F_B$ are mapped using an inverse function $\psi^{-1}$, resulting in the set $G^{\prime}$ (Line~\ref{line:map_features}). 
The algorithm then iterates through each candidate object $P \in D^C\cup D^I$, projects its properties onto the selected subset $F_B$ to obtain the property vector $G_{P_{BK}}$, and adds it to the candidate set $D^C_{BK}$ and index set $D^I_{BK}$ (Lines~\ref{line:extract_P}–\ref{line:add_to_D_C} and~\ref{line:extract_P_index}-\ref{line:add_to_D_I}, respectively). The index set $D^I_{BK}$ is then indexed and stored in $BKAFI_{Ind}$ (Line~\ref{line:index_D_I}). The algorithm continues by iterating through each modified feature vector $G_{P_{BK}}$ in $D^C_{BK}$ (Lines~\ref{line:map_back_P}-\ref{line:add_to_C}). For each vector, the corresponding object $P$ is mapped back from the feature vector to its item in $D^C$ and the $k$ nearest neighbors are retrieved from the indexed set $BKAFI_{Ind}$ (Line~\ref{line:retrieve_neighbors}). The algorithm then maps each nearest neighbor back to the corresponding item in $D^I$ and adds the candidate pair $(P, P_I)$ to the set $C$ (Line~\ref{line:add_to_C}). Finally, the set of candidate pairs $C$ is returned as the output of the algorithm (Line~\ref{line:return_C}).

We use a \emph{KDTree} index~\cite{bentley1975multidimensional, ram2019revisiting}, a proven efficient and effective indexing method for low-dimensional spaces, to index the set $D^I$ in the $|F_B|$-dimensional feature space. Its construction (Line~\ref{line:index_D_I}) takes $O(|D^I| \log |D^I|)$~\cite{bentley1975multidimensional}, which is a one-time cost and typically negligible compared to the cost of performing nearest neighbor searches for all candidates in $D^C$. Under the assumption of low dimensionality, each query takes $O(\log |D^I|)$, making the method scalable for large $|D^C|$. While the parameter $k$ has little impact on the query complexity, it directly impacts the number of candidate pairs generated per item, which in turn affects the cost of the downstream matching phase. 

\revB{To mitigate pair over-generation, we refrain from returning exactly $k$ nearest neighbors. Instead, we incorporate a retrieval-time similarity-based pruning, allowing early process termination whenever the distance in the sub-property feature space is above a calibrated threshold, derived from training data. This adjustment reduces false positives and supports more efficient matching without compromising recall. 
Specifically, we compute distances between matching pairs in the training set and set a high threshold, \emph{e.g.}, $95$-th percentile of the distribution. At inference time, retrieval of candidates proceeds in increasing distance order  until either $k$ neighbors are collected or the distance to the next neighbor exceeds the threshold. It is worth noting that a higher quantile increases robustness to noise but may miss potential matches, while a lower quantile favors recall at the risk of introducing false positives.} 

As for feature set size $|F_B|$, we empirically observe (Section~\ref{subsection:blocking_bkafi_dim_experiments}) that smaller subsets of informative features, selected in a data-driven manner, yield faster and more effective search.

\subsubsection{Matching}
\label{subsection:matching}
The matching process is performed using the candidate set $C$, which was obtained in the blocking phase. The model $\tilde{\theta}$ (Section~\ref{subsection:train}), is then used for prediction. For each pair in $C$, we run the featurization process (Section~\ref{subsubsection:property_vectors}), and subsequently pass it through the trained model to generate a match/ non-match decision.

%% file: hague_benchmark.tex
\section{New Benchmark for 3D Geospatial ER}
\label{sec:hague_benchmark}
To the best of our knowledge, \fulltaskspace has not been previously studied, and no standardized benchmarks exist in the literature. To address this gap, we curated a new benchmark dataset from the city of \emph{The Hague}, constructed from two distinct sources, enabling a natural partition into candidate and indexed sets. Below, we detail the dataset construction process and describe its different variants. Key statistics are summarized in Table~\ref{tab:hague_datasets}.

\noindent\textbf{Data Sources:} The 3D City Model \emph{The Hague 2022} (the candidate set) is maintained by the municipality of \emph{The Hague} as part of the Open Data platform.\footnote{\url{https://denhaag.dataplatform.nl/}} It is derived from 2021 aerial photographs and structured according to the Register of Buildings and Addresses (\emph{BAG}). The raw data comprises 46 CityJson files, each describing a distinct district, where the coordinate system used in all files is RDnew, the national coordinate system of the Netherlands.

\emph{3DBAG} (the index set) is a large-scale openly available 3D building dataset, containing over 10 Million buildings across the Netherlands~\cite{peters2022automated}. The data was automatically generated from three sources of the Topographic Register of the Netherlands -- \emph{BAG},\footnote{\url{https://www.kadaster.nl/zakelijk/registraties/basisregistraties/bag}} the National Height Model of the Netherlands (AHN),\footnote{\url{https://www.ahn.nl/}} and \emph{TOP10NL}\footnote{\url{https://www.kadaster.nl/zakelijk/producten/geo-informatie/topnl}} -- using various methods such as aerial photography, terrestrial surveying and airborne laser scanning.
\revC{The dataset is publicly available for download\footnote{\url{https://3dbag.nl/en/download}} and is partitioned into spatial tiles for efficient access. Since our focus is on \emph{The Hague}, we selected 246 tiles covering more than 100,000 buildings within the city’s boundaries, providing spatial overlap with the first data source.}

Both data sources include unique object identifiers, enabling a direct mapping between them after standardizing key formats to resolve format inconsistencies. To guarantee data quality and consistency, we retained only objects (buildings) composed of at least 10 polygons. This threshold eliminates incomplete or overly simplified structures, ensuring that the retained objects exhibit sufficient geometric complexity for meaningful evaluation. 

\begin{table}[t]
\centering
\caption{\revC{Statistics of training and test sets averaged over three random seeds. Candidate and index sizes are shown only for test sets, as they are relevant to the blocking evaluation. The training set is used solely for learning the matcher.}}
\begin{tabular}{llrrr}
\toprule
\textbf{Set} & \textbf{Size} & \textbf{Candidates} & \textbf{Index} & \textbf{Total Pairs} \\
\midrule
Train  & Small  & --     & --     & 14,178 \\
       & Large  & --     & --     & \revC{60,379} \\
\midrule
Test   & Small  & 3,507  & 9,985  & 13,789 \\
       & Large  & \revC{15,588}  & \revC{97,437} & \revC{52,586} \\
\bottomrule
\end{tabular}
\label{tab:hague_datasets}
\end{table}

\vspace{.1in}
\noindent\textbf{Dataset Variants:} To ensure robust evaluation, we generate two dataset variants by varying the training and testing configurations across three random seeds. Specifically, we sample training sets of the following sizes: \emph{small} ($0.1$) and \emph{large} ($0.6$). These ratios are applied to the intersection of candidate and index object identifiers to determine the number of matching training samples.

For training, we construct different pair sets for the blocking and matching components. For blocking we enforce that every object in the candidate set has a corresponding match in the index set
by retaining, in the candidate set, only objects that appear in both sources. For each such candidate object, its corresponding index object is included to form a positive pair. We apply negative sampling 
by generating 
two non-matching pairs for each matching pair, selected at random.
For matching, to yield a challenging and realistic evaluation scenario, we use the output of the best-performing blocking configuration (see Section~\ref{sec:blocking_experiments}). Specifically, we generate a candidate pair set by retrieving the top-3 nearest neighbors for each candidate object in the property space, with the true match always included, even if it is not among the top-3 results.

To construct the test data, we follow a two-step strategy over both blocking and matching. For blocking, we generate candidate and index sets by identifying test candidate objects from the two source intersection, excluding any identifiers used in the training data. For each test size (\emph{small} ($0.1$) and \emph{large} ($1.0$)) we randomly sample a subset of remaining candidates, where the ratio in parentheses indicates the fraction of the remaining candidate pool used. \revC{To evaluate BKAFI's reduction power we augment the candidate set so that $20\%$ of candidates do not have a matching object in the index set.} The corresponding index set is constructed by 
copying the sampled test candidate set, then augmenting it with random samples from the full index object pool. The size of this additional sample set is determined by applying the same test ratio to the total number of index objects. 
For 
matching, 
we 
sample from the unused portion of the intersection between the two sources, excluding any object identifiers that were present in the training pairs. For each test size, we sample a set of matching test pairs by selecting candidate objects and pairing them with their corresponding index matches. 
To form a set of non-matching pairs, we again use the best-performing blocking configuration to retrieve the top-3 nearest index objects for each candidate in the sampled test set. These top-3 results form the set of candidate pairs for evaluation, ensuring the inclusion of the true match. This design ensures a realistic yet challenging evaluation scenario for the matcher.

The reported values (Table~\ref{tab:hague_datasets}) represent averages computed over the three different random seeds employed during data partitioning. To the best of our knowledge, this is the first work to leverage polygon meshes for \task, introducing a new modality to the domain. \revA{To facilitate further research the dataset is released as a benchmark for \fulltask.\footnote{\url{https://tinyurl.com/3dSAGERdataset}}}

%% file: experiments_new.tex
\section{Empirical Evaluation}\label{sec:experiments_full}
Equipped with the new benchmark
, we now evaluate \model. Section~\ref{sec:experimental_setup} describes the experimental setup, followed by blocking and matching evaluation (sections~\ref{sec:blocking_experiments} and~\ref{sec:matching_evaluation}, respectively).

\input{experimental_setup}

\input{blocking_experiments}
\input{matching_experiments}

%% file: experimental_setup.tex
\subsection{Experimental Setup}
\label{sec:experimental_setup}
We outline the experimental framework, starting with a description of an additional dataset \revGeneral{used for our case study (Section~\ref{subsection:proprietary_dataset}).} We then provide implementation details (Section~\ref{subsection:settings}), describe the evaluation metrics (Section~\ref{subsection:evaluation}), present the baselines (Section~\ref{subsection:baselines}) and conclude with key design choices (Section~\ref{subsection:hyperparameters}).

\subsubsection{\revGeneral{Case Study Data}}
\label{subsection:proprietary_dataset}
\revGeneral{In addition to our new benchmark (Section~\ref{sec:hague_benchmark}), we collaborated with a third-party partner to obtain real-world data from a post-disaster scenario. This data is proprietary and thus cannot be released publicly. It was collected from two distinct sources and consists of 3D objects represented in LOD2. The data was generated from point clouds and approximate building footprints, with roofs reconstructed using plane adjustment techniques. We retained only objects with at least $10$ polygons, and applied the same partitioning policy as introduced in Section~\ref{sec:hague_benchmark}, resulting in 5,472 and 3,705 total pairs for Train and Test, respectively. For the blocking component analysis, the test set is split into 938 candidate objects and 9,617 index objects. We refer to this dataset as PDM (Post-Disaster Management).}

\subsubsection{Implementation Details}
\label{subsection:settings}
All experiments were conducted on a server running CentOS 7, equipped with 2 Nvidia Quadro RTX 6000 GPUs. To ensure robust evaluation, each experiment was repeated across three random initialization seeds. Our implementation is publicly available, and the full list of features used in our experiments can be found in the repository configuration file.

\sloppy
To instantiate the model $\tilde{\theta}$ (see Section~\ref{subsection:train}), we evaluated a variety of machine learning algorithms on the \emph{Hague\textsubscript{small}} dataset, including \texttt{Random\allowbreak Forest} (RF)~\cite{breiman2001random}, \texttt{Ada\allowbreak Boost} (AB)~\cite{freund1996experiments}, \texttt{Gradient\allowbreak Boosting} (GB)~\cite{friedman2001greedy}, \texttt{Bagging} (B)~\cite{breiman1996bagging}, and \texttt{Extreme\allowbreak Gradient\allowbreak Boosting} (XGB)~\cite{chen2016xgboost}. We also evaluated a neural baseline implemented as a multilayer perceptron (MLP) with 2 or 3 fully connected layers.
The XGBoost model uses the official open-source implementation,\footnote{\url{https://xgboost.readthedocs.io/}} while all other models rely on the Scikit-learn library~\cite{pedregosa2011scikit}.\footnote{\url{https://scikit-learn.org/}} Each model was trained using grid search with 5-fold cross-validation over a predefined hyperparameter space, with configuration details provided in our GitHub repository. The \texttt{Bagging} model performance dominated the alternatives (see Section~\ref{subsection:exp_matching_backbone_model_comparison}) and was adopted as the matcher of choice for the evaluation.

\subsubsection{Evaluation Metrics}
\label{subsection:evaluation}
\sloppy
\revB{In line with \revC{prior work~\cite{paulsen2023sparkly}}, we evaluate blocking effectiveness considering the tradeoff between match coverage and pruning efficiency. The former is measured by \emph{Pair Completeness} (PC), the proportion of matching pairs retained after the blocking step (corresponds to recall). The second is \emph{Reduction Ratio} (RR), the reduction in the number of pairwise comparisons relative to the full cross-product of the candidate and index sets~\cite{papadakis2019survey}. Unless stated otherwise, RR is governed by the parameter $k$, set to a maximum of $20$, without applying the distance threshold-based pruning (Section~\ref{subsection:blocking}). This value reflects realistic candidate list sizes commonly used in downstream matching tasks~\cite{mudgal2018deep,li2021deep}. 
Efficiency is evaluated in terms of execution time (in seconds).}

For matching, in line with prior work~\cite{christen2012data,li2021deep}, we evaluate performance using precision, recall and F1-score. We also analyze scalability using runtime and memory usage. This includes measuring training and inference time, as well as model size, offering a more comprehensive view of deployment feasibility.

\revGeneral{For the case study (PDM dataset, see Section~\ref{subsection:proprietary_dataset}), we focus on a specific measurable KPI, \textbf{Mutual Registration Accuracy} (MRA)~\cite{tam2012registration}, which can be derived from Recall@K.}

\subsubsection{Baselines}
\label{subsection:baselines}
Since \fulltaskspace has not been previously addressed in the literature, there are no established baselines for direct comparison. To enable meaningful evaluation, we adapt methods from related tasks and construct competitive alternatives to benchmark the performance of \model.

\begin{figure*}[t]
    \centering
    \begin{subfigure}[t]{0.325\textwidth}
        \centering
        \includegraphics[width=\textwidth]{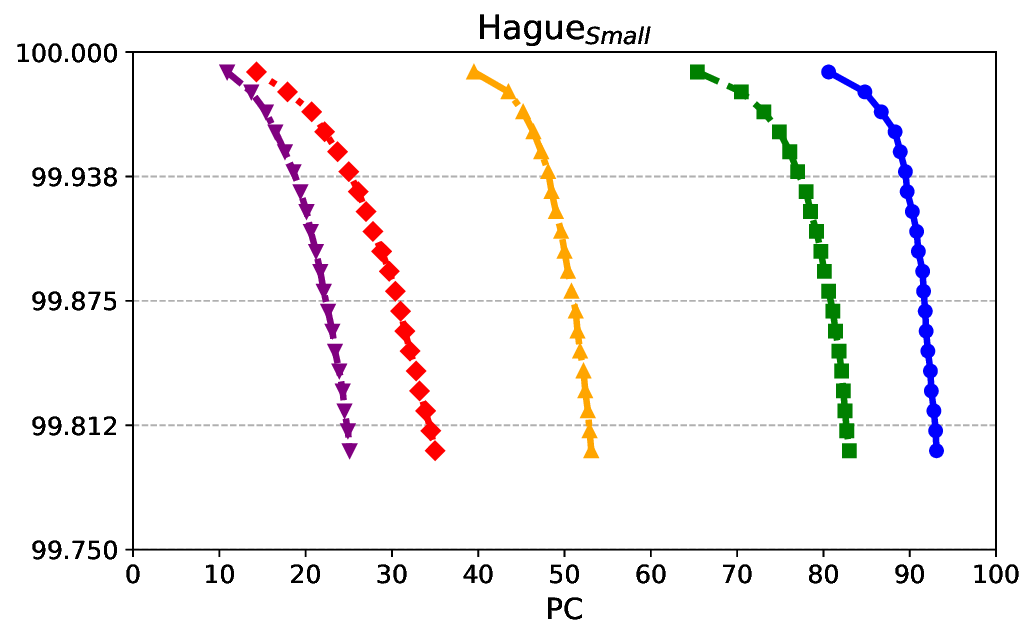}
        \caption{\emph{Hague\textsubscript{small}}}
        \label{fig:blocking_rr_vs_recall_hague_small}
    \end{subfigure}
    \hfill
    \begin{subfigure}[t]{0.335\textwidth}
        \centering
        \includegraphics[width=\textwidth]{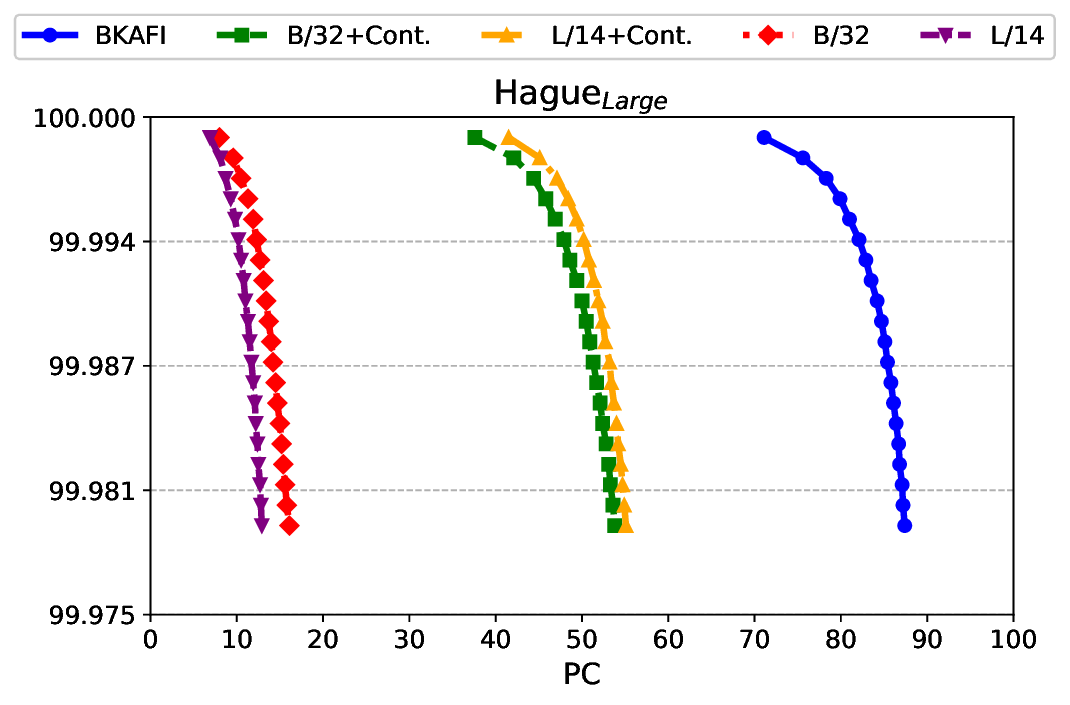}
        \caption{\emph{Hague\textsubscript{large}}}
        \label{fig:blocking_rr_vs_recall_hague_large}
    \end{subfigure}
    \hfill
    \begin{subfigure}[t]{0.325\textwidth}
        \centering
        \includegraphics[width=\textwidth]{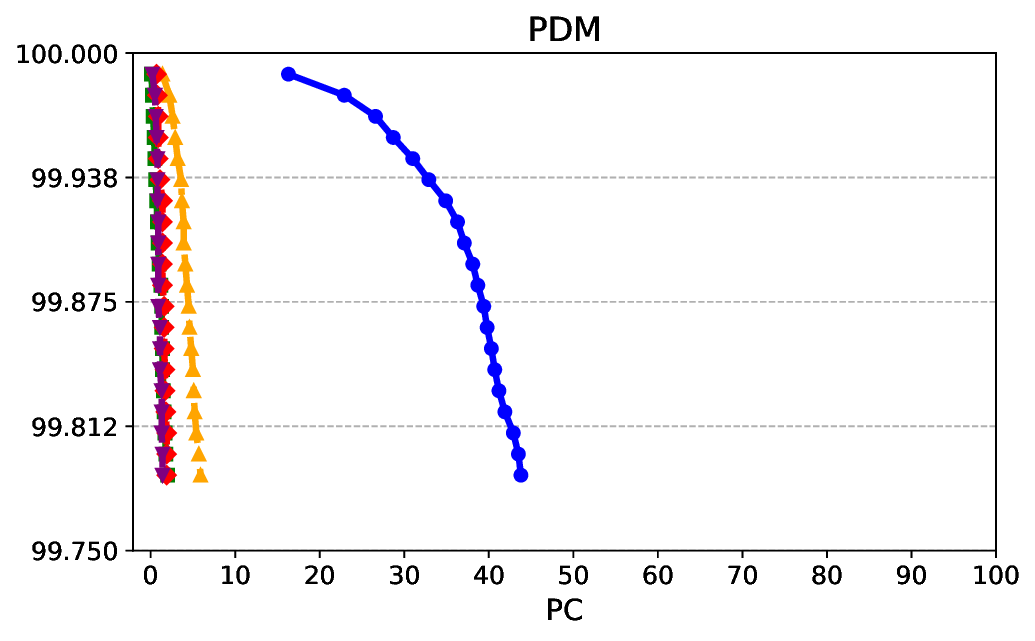}
        \caption{\revGeneral{\emph{PDM dataset}}}
        \label{fig:blocking_rr_vs_recall_PDM}
    \end{subfigure}
    \caption{RR vs. PC comparison between BKAFI and baseline methods across the different datasets.}
    \label{fig:blocking_RR_vs_PC_baselines_all}
\end{figure*}

\noindent\textbf{\emph{Blocking Baselines}:}
\label{subsection:blocking_baselines}
We evaluate the following baselines:

\noindent $\bullet$ \textbf{ViT}: Visual representations of \mesh es are generated by rendering objects as PNG images (see Figure~\ref{fig:matching_example}). Embedding vectors are extracted using pretrained Vision Transformer (ViT) models~\cite{dosovitskiy2020vit}, followed by similarity search using Faiss~\cite{johnson2019billion} to retrieve potential matches. 
We experiment with two variants: \emph{ViT-B/32} and \emph{ViT-L/14}.
This approach tests whether image-based representations effectively captures structural similarity between buildings, offering a blocking strategy that is independent of explicit spatial information.

\noindent $\bullet$ \textbf{ViT+Contrastive}: This baseline builds on the ViT setup by fine-tuning the visual encoder using contrastive learning. We use training set pairs (as described in Section~\ref{sec:hague_benchmark}), where each pair consists of a candidate and an index object rendered as PNG images. The model architecture reuses the visual encoder from a pretrained Vision Transformer (ViT), and is trained with a contrastive loss (Cosine Embedding Loss) to pull matching pairs closer in the embedding space while pushing apart non-matching ones. We use the CLIP implementation of both \emph{ViT-B/32} and \emph{ViT-L/14} as the backbone and train for 8 and 5 epochs per random seed, respectively, after observing that the contrastive loss converges within these ranges. After training, the encoder is used to extract refined embeddings for similarity-based blocking with Faiss.

\noindent\textbf{\emph{Matching Baselines}:}
\label{subsection:matching_baselines}
We use the following baselines: 

\noindent $\bullet$ \textbf{ViT+Contrastive}: Similar to the blocking setup, we fine-tune a visual encoder using contrastive learning to bring matching object pairs closer in the embedding space. However, in this case, the goal is to perform binary classification (match or non-match). Specifically, we build a lightweight binary classifier on top of a frozen visual encoder (e.g., \texttt{ViT-B/32} or \texttt{ViT-L/14}) from the CLIP model~\cite{radford2021learning}. The classifier, implemented as a two-layer feedforward network with ReLU activation, receives the absolute difference between the image embeddings as input and learns to predict the matching decision boundary.
The model is trained using binary cross-entropy loss (\texttt{BCEWithLogitsLoss}) over labeled image pairs, with training data constructed from matching and non-matching object pairs (see Section~\ref{sec:hague_benchmark}). Training is conducted using the AdamW optimizer with a learning rate of $1\text{e}{-5}$ and a batch size of 4.

\noindent $\bullet$ \textbf{LLM (GPT-4o)}: Given the predominant performance of GPT-4o over other LLMs in traditional ER problems~\cite{peeters2025entity}, we implemented an LLM-baseline for \fulltask. Given the supervised learning setup, we provided examples in the prompt (in-context learning). The performance of GPT-4o was inferior and thus omitted from the results. This may be attributed to the model’s inability to effectively process raw polygon mesh representations.

\subsubsection{Hyperparameters and Design Choices}
\label{subsection:hyperparameters}
For BKAFI, aside from the detailed component analysis in Section~\ref{subsection:blocking_bkafi_dim_experiments}, we report results using the value of $|F_{B}|$ (i.e., the number of selected blocking key functions) that achieved the best area under the PC@k curve (AUC) on the training set.  In addition to comparing different model architectures for matching, we employ the \texttt{Bagging} model for the ablation study. Feature importance scores computation, used as a preparatory step for blocking, are performed using a \texttt{Random\allowbreak Forest} model, chosen for its efficiency and fast training time.

%% file: blocking_experiments.tex
\subsection{Blocking Evaluation}
\label{sec:blocking_experiments}
This section presents the experimental analysis of BKAFI, our blocking approach. We begin by comparing BKAFI with baseline methods (Section~\ref{subsection:blocking_bkafi_vs_baselines}), followed by an ablation study analyzing its key components, namely 
normalization (Section~\ref{subsection:normalization_ablation}), and the property selection criterion used to define the blocking keys (Section~\ref{subsection:blocking_keys_criterion}). We also analyze the effect of $|F_{B}|$, the number of properties dictating the search space size \revC{and Cardinality, in Section~\ref{subsection:blocking_bkafi_dim_experiments}}. \revGeneral{Finally, we provide a case-study analysis over the PDM dataset (Section~\ref{subsection:mra}).}

\subsubsection{BKAFI vs. Baselines}
\label{subsection:blocking_bkafi_vs_baselines}
We start by comparing BKAFI against the baseline methods (Section~\ref{subsection:blocking_baselines}), along the dimensions of effectiveness and runtime. 
Our analysis shows that BKAFI outperforms all baselines over all experiment settings and does that 1-2 order of magnitudes faster.

\noindent \textbf{Effectiveness}:
\label{subsubsection:rr_vs_recall_against_baselines}
Figure~\ref{fig:blocking_RR_vs_PC_baselines_all} compares BKAFI with the four baseline models in terms of RR versus PC, with both metrics reported as percentages. Each RR value corresponds to a specific $k$ used in the candidate generation step. For example, for \emph{Hague\textsubscript{small}}, setting $k=20$ yields an RR of $100 \left(1 - \frac{20 \times 3{,}507}{3{,}507 \times 9{,}985}\right) \approx 99.7997$. Intuitively, being farther to the right on the plot is better, as it indicates higher PC score for a given RR value.

\revGeneral{
We first observe that BKAFI dominates the baselines for all datasets. For example, over \emph{Hague\textsubscript{large}} BKAFI achieves $PC=81\%$ for $k=5$ ($RR \approx 99.9949\%$), compared to $46.9\%$ and $49.4\%$ contrastive learning-based baselines, \texttt{B/32+Cont.} and \texttt{L/14+Cont.}, respectively. All methods experience a drop in PC while moving from \emph{Hague\textsubscript{small}} to \emph{Hague\textsubscript{large}}. This is expected, as a larger index set ($97,437$ objects compared to $9,985$) inherently contain more potential options per object from the candidate set, making the blocking task more challenging. Interestingly, the performance degradation is more pronounced for baseline methods, particularly the raw ViT-based models, indicating their limited scalability to larger search spaces. In contrast, BKAFI maintains robust performance across dataset sizes, with only moderate declines in PC, especially in the low-$k$ range. This suggests that BKAFI is better equipped to handle increased data complexity and scale.}

\revGeneral{On the \emph{PDM} dataset, which contains simpler 
geomteries and is thus more challenging for image-based similarity alone, the gap widens further. 
BKAFI reaches $PC=31.2\%$ at $RR \approx 99.948\%$ ($k=5$), whereas the best baseline, \texttt{L/14+Cont.}, only achieves $3.2\%$, and others 
below $2\%$. This sharp contrast highlights the advantage of BKAFI's geometry-aware blocking mechanism, particularly where visual embeddings struggle to capture fine-grained structural cues.}

Results demonstrate that BKAFI not only scales well with dataset size but also generalizes better to diverse 3D modeling conditions, making it a reliable choice for handling \fulltask.

\label{subsubsection:execution_time_experiments}

\begin{table}[b]
\centering
\caption{\revGeneral{Blocking Execution Time (in seconds)}} 
\begin{tabular}{lrrr}
\toprule
\textbf{Method} & \textbf{Hague\textsubscript{small}} & \textbf{Hague\textsubscript{large}} & \revGeneral{\textbf{PDM}} \\
\midrule
BKAFI           & {\bf 0.22}    & {\bf 1.08}    & {\bf 0.08} \\
\midrule\midrule
B/32+Cont.      & 5.09    & 279.43   & 1.29 \\
L/14+Cont.      & 9.22    & 421.29   & 2.50 \\
B/32            & 2.69    & 176.75   & 1.30 \\
L/14            & 4.91    & 264.77   & 2.58 \\
\bottomrule
\end{tabular}
\label{tab:execution_times}
\end{table}

\noindent\textbf{Execution Time:}
Table~\ref{tab:execution_times} provides a runtime comparison of all methods across different dataset sizes. BKAFI is substantially faster than baselines. For example, on the \emph{Hague\textsubscript{large}} dataset, BKAFI completes the blocking step in just \revGeneral{$1.08$} seconds, compared to \revGeneral{$279.43$} and \revGeneral{$421.29$} seconds for the \texttt{B/32+Cont.} and \texttt{L/14+Cont. } baselines, respectively, amounting to over two orders of magnitude speedup.

BKAFI's speedup may be attributed to its compact feature representation (Section~\ref{subsection:blocking}). While the baseline models operate in high-dimensional embedding spaces \revGeneral{(512 and 768 for B/32 and L/14, respectively)}, BKAFI uses only the most important features as a blocking key. Also, ViT-based baselines incur overhead due to the exhaustive embedding generation for all objects, a step involving forward passes through a heavy neural backbone. For the fine-tuned variants, the cost is further amplified due to a separate training phase over image pairs. Image rendering, a non-negligible one-time preprocessing step, is excluded from the reported runtime. The comparative analysis underscores BKAFI’s suitability for large-scale deployments, where efficiency and scalability are critical.

\subsubsection{Effect of Normalization}
\label{subsection:normalization_ablation}
We now asses property value normalization impact (Section~\ref{subsubsection:property_vectors}). For ease of presentation, we report PC@$k$ ($k \in \{1, 5, 20\}$) instead of RR, as it offers a direct view of recall at fixed candidate list sizes. As shown in Table~\ref{tab:normalization_effect}, normalization consistently improves PC@$k$ across all datasets, particularly at lower $k$ values. For instance, on \emph{Hague\textsubscript{small}}, PC@$1$ improves from \revGeneral{$29.5\%$ to $80.6\%$}, and on \emph{Hague\textsubscript{large}}, from \revGeneral{$59.2\%$ to $71.1\%$}. Gains at $k=20$ are smaller but still present. \revGeneral{On the \emph{PDM} dataset, although the absolute PC values are lower, normalization still leads to a measurable gain. For example, PC@$5$ increases from $28.8\%$ to $31.0\%$. These results suggest that normalization is a simple yet effective preprocessing step for enhancing property similarity-based blocking.}

\begin{table}[b]
\centering
\caption{\revGeneral{PC@$k$ with and without normalization.}}
\begin{tabular}{lcccc}
\toprule
\textbf{Dataset} & \textbf{Normalization} & \textbf{PC@1} & \textbf{PC@5} & \textbf{PC@20} \\
\midrule
\multirow{2}{*}{Hague\textsubscript{small}} 
  & Yes & 80.6 & 88.9 & 93.1\\
  & No  & 29.5 & 67.4 & 88.1 \\
\midrule
\multirow{2}{*}{Hague\textsubscript{large}} 
  & Yes & 71.1 & 81.0 & 87.4 \\
  & No  & 59.2 & 78.1 & 86.3 \\
\midrule
\multirow{2}{*}{\revGeneral{PDM}} 
  & Yes & 16.3 & 31.0 & 43.8 \\
  & No  & 14.6 & 28.8 & 41.2 \\
\midrule
\bottomrule
\end{tabular}
\label{tab:normalization_effect}
\end{table}

\subsubsection{Blocking Key Selection Criterion}
\label{subsection:blocking_keys_criterion}
An essential component of BKAFI is the selection of blocking key functions. This step determines the subset of features $F_B$ used to construct the similarity space for nearest neighbor retrieval.
We propose to use the top-$|F_B|$ features according to their importance scores extracted from the trained model $\tilde{\theta}$. We also explore a statistical selection strategy, where features are ranked based on the standard deviation of their corresponding property ratios across matching pairs, in ascending order. Intuitively, features with lower standard deviation, indicating more stable ratios between matched objects, might be more informative for identifying true matches.

\begin{table}[t]
\centering
\caption{\revGeneral{PC@$k$ over different blocking key criteria.}}
\begin{tabular}{lcccc}
\toprule
\textbf{Dataset} & \textbf{Selection Criterion} & \textbf{PC@1} & \textbf{PC@5} & \textbf{PC@20} \\
\midrule
\multirow{2}{*}{Hague\textsubscript{small}} 
  & Feature Importance & 80.6 & 88.9 & 93.1 \\
  & Std                & 40.2 & 64.3 & 80.6 \\
\midrule
\multirow{2}{*}{Hague\textsubscript{large}} 
  & Feature Importance & 71.1 & 81.0 & 87.4 \\ 
  & Std                & 29.9 & 48.6 & 66.1 \\
\midrule
\multirow{2}{*}{\revGeneral{PDM}} 
  & Feature Importance & 16.3 & 31.0 & 43.8 \\
  & Std                & 11.4 & 22.1 & 35.3 \\

\bottomrule
\end{tabular}
\label{tab:blocking_key_selection}
\end{table}

\revGeneral{Table~\ref{tab:blocking_key_selection} presents PC@$k$ of both selection strategies. For \emph{Hague\textsubscript{small}}, feature importance consistently outperforms standard deviation across all $k$ values. This dominance is even more evident for \emph{Hague\textsubscript{large}}, particularly at lower $k$: PC@$1$ improves from $29.9\%$ (standard deviation) to $71.1\%$ (feature importance). A similar trend appears on the \emph{PDM} dataset, where PC@$1$ improves from $11.4\%$ to $16.3\%$, and PC@$5$ from $22.1\%$ to $31.0\%$. While standard deviation may serve as a lightweight, training-free proxy for feature selection, especially in smaller datasets, feature importance provides robust performance, particularly in larger and complex settings.}

\subsubsection{Effect of Search Space Size ($|F_B|$)}
\label{subsection:blocking_bkafi_dim_experiments}
A key parameter of BKAFI is the number of selected blocking key features ($|F_B|$). This value determines the dimensionality of the property space in which nearest neighbor retrieval is performed. At runtime, only the top $|F_B|$ properties are used to construct the vector representations of candidate and index objects. 

\begin{figure}[b]
    \centering
    \includegraphics[width=0.96\columnwidth]{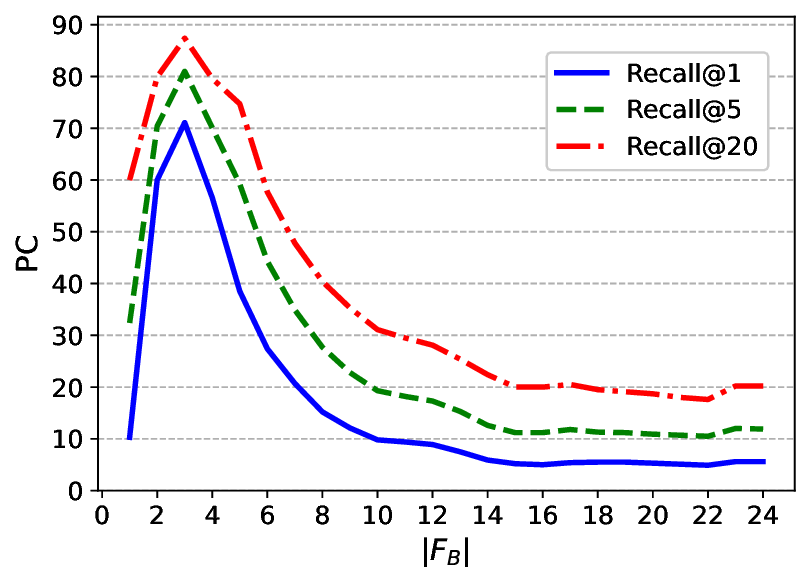}
    \caption{PC@$k$ vs. search space size $|F_B|$ for $k\in\{1,5,20\}$ across Hague\textsubscript{large}.}
    \label{fig:PC_vs_fB_large}
\end{figure}

\noindent\textbf{PC@k vs. Search Space Size ($|F_B|$):}
\label{subsubsection:PC_vs_fB_large}
Figure~\ref{fig:PC_vs_fB_large} shows how PC@$k$ varies with $|F_B|$ for $k\in\{1,5,20\}$ over \emph{Hague\textsubscript{large}}. We observe a steep increase in PC as $|F_B|$ grows from 1 to 2--4, with performance peaking early. In all cases, the highest PC is achieved using a compact feature subset ($|F_B|=3$), after which PC steadily declines and stabalizes at $|F_B|\approx 14$.

This observed trend can be attributed to the nature of the blocking strategy: while incorporating a small number of discriminative properties improves candidate filtering, adding too many properties leads to over-restriction of the search space. As the size of a blocking key increases, fewer candidate pairs survive the filter, resulting in PC loss. This is especially pronounced for PC@1, where early elimination of good candidates becomes more detrimental.
These results highlight the importance of a compact and selective property space in blocking. An aggressive property expansion harms PC due to reduced overlap, reinforcing the value of principled feature selection in the BKAFI framework.

\noindent\textbf{Execution Time vs. Search Space Size ($|F_B|$) and Cardinality:}
\label{subsubsection:runtime_vs_dim}
\revC{Figure~\ref{fig:runtime_vs_fB} presents the execution time as a function of the dataset cardinality (the cross product of he candidate and index set), for three different $|F_B|$ values. The cardinality values shown on the X-axis refer to four variants of \emph{The Hague} dataset:  the previously defined \textit{small} and \textit{large} versions (see Table~\ref{tab:hague_datasets}), as well as two additional variants created specifically for this analysis. Each subplot corresponds to a different value of $|F_B| \in \{3, 15, 24\}$, which dictates the dimensionality of the sub-property vectors.}

\begin{figure}[b]
    \centering
    \includegraphics[width=0.993\columnwidth]{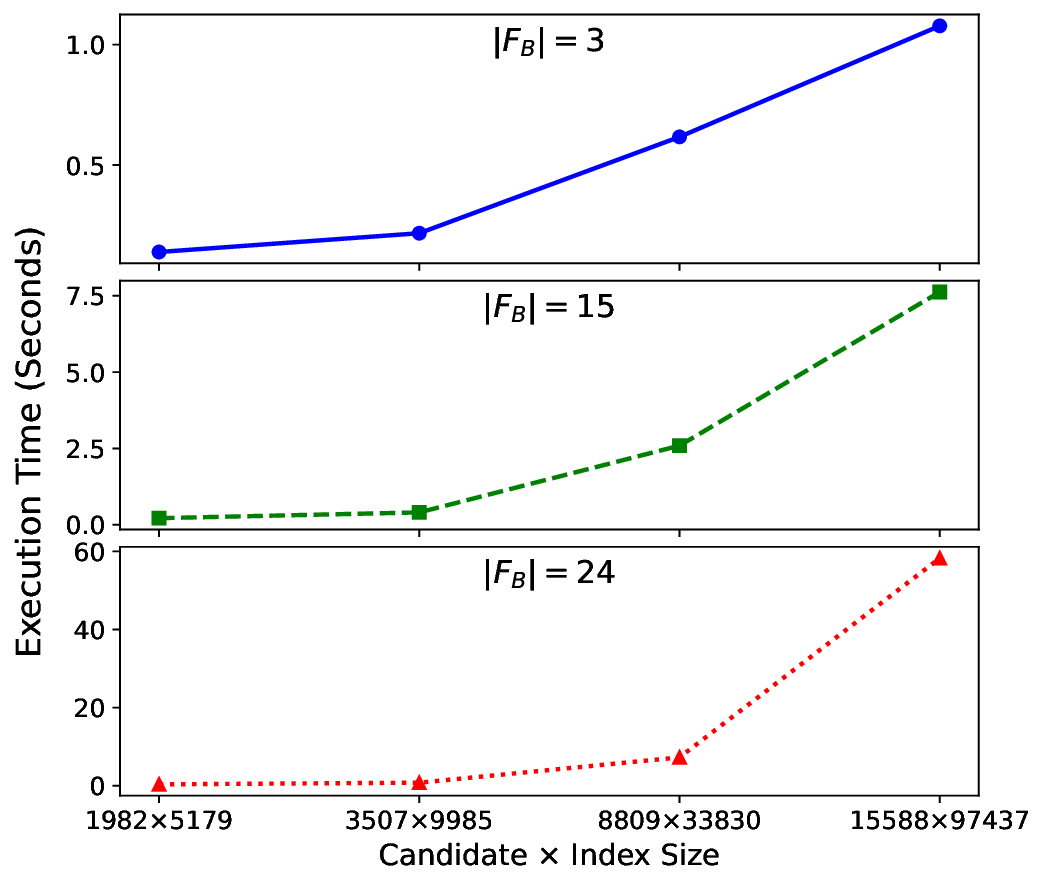}
    \caption{\revC{Execution time vs. cardinality of \emph{The Hague} dataset variants, with different $|F_B|$ values. }}
    \label{fig:runtime_vs_fB}
\end{figure}

\revC{We observe that runtime increases with both data volume and feature dimensionality, as expected. However, the rate of increase is moderate for lower-dimensional setups (e.g., $|F_B|=3$), where even large-scale blocking remains computationally efficient. For the $|F_B|=24$ setting shows a steep rise in execution time for the largest dataset, exceeding 50 seconds. These findings reinforce that 
BKAFI 
efficiently scales when operating in compact subspaces, a design supported by our earlier results showing that high recall can be achieved using a small number of features (see Figure~\ref{fig:PC_vs_fB_large}).}

\ifdefined\TechReport
Figure~\ref{fig:runtime_vs_fB} presents a more detailed analysis over all possible $|F_B|$ values, for the small and large variants of \emph{The Hague} dataset. Specifically As expected, runtime generally increases with $|F_B|$, reflecting the added cost of higher-dimensional similarity search and candidate filtering. This growth is most prominent for the large variant, where execution time rises sharply as $|F_B|$ increases. The observed trend reinforces the conclusions from Figure~\ref{fig:PC_vs_fB_large}: near-optimal PC is achieved with a small number of features, which also yields faster runtime. That means, BKAFI benefits from compact feature spaces both in accuracy and scalability.

\begin{figure}[t]
    \centering
    \includegraphics[width=0.98\columnwidth]{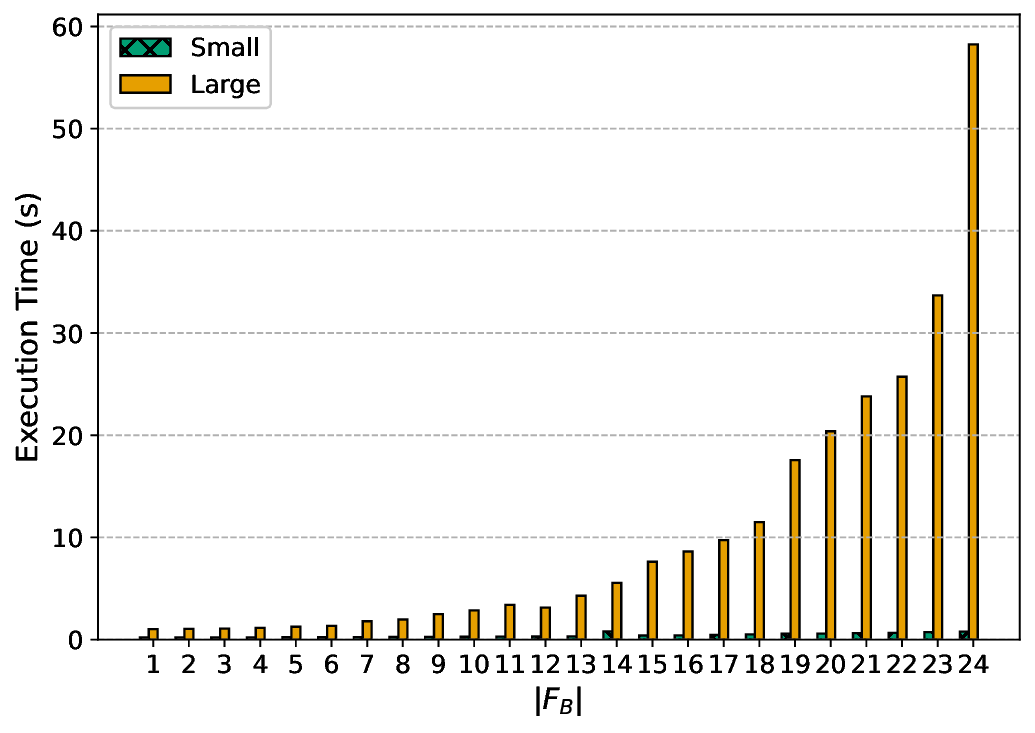}
    \caption{\revGeneral{Execution time vs. $|F_B|$ for \emph{The Hague} datasets.}}
    \label{fig:runtime_vs_fB}
\end{figure}

We note that the plot exhibits occasional irregularities, \emph{e.g.}, small dips or spikes in execution time for certain $|F_B|$ values. These fluctuations likely stem from implementation-level effects such as system caching, multithreading variability, or how the underlying similarity structures are partitioned in memory. Despite these, the overall trend remains clear: smaller $|F_B|$ leads to significantly more efficient blocking.

To summarize, our empirical analysis shows that low $|F_B|$ values not only achieve higher recall, but also lead to lower search times, making them preferable both in terms of effectiveness and efficiency.
\fi

\subsubsection{\revGeneral{Mutual Registration Accuracy for Post Disaster Management:}}
\label{subsection:mra}
\revGeneral{In post-disaster scenarios such as the one captured by the PDM dataset (Section~\ref{subsection:proprietary_dataset}), decision-making must account for strict resource constraints as computational resources are often limited due to urgency, infrastructure damage, or cost considerations~\cite{zhuo2017automatic,manzini2024crasar}. Therefore, data-driven prioritization is critical. The PDM literature uses registration-based analysis~\cite{tam2012registration} and, in particular, mutual registration accuracy (MRA), as a decision-making tool. Basically, MRA reports on the success of identifying same real-world entities in a symmetric manner, both from the indexed dataset to the candidate dataset and vice versa, given limited matching resources. In a clean-clean setting (see Section~\ref{subsubsection:contamination_experiments} for the impact of contamination on the analysis), where $D^C\subset D^I$, MRA under budget constraints is translated into PC (see Section~\ref{subsection:evaluation}).}

\begin{figure}[t]
    \centering
    \includegraphics[width=0.999\columnwidth]{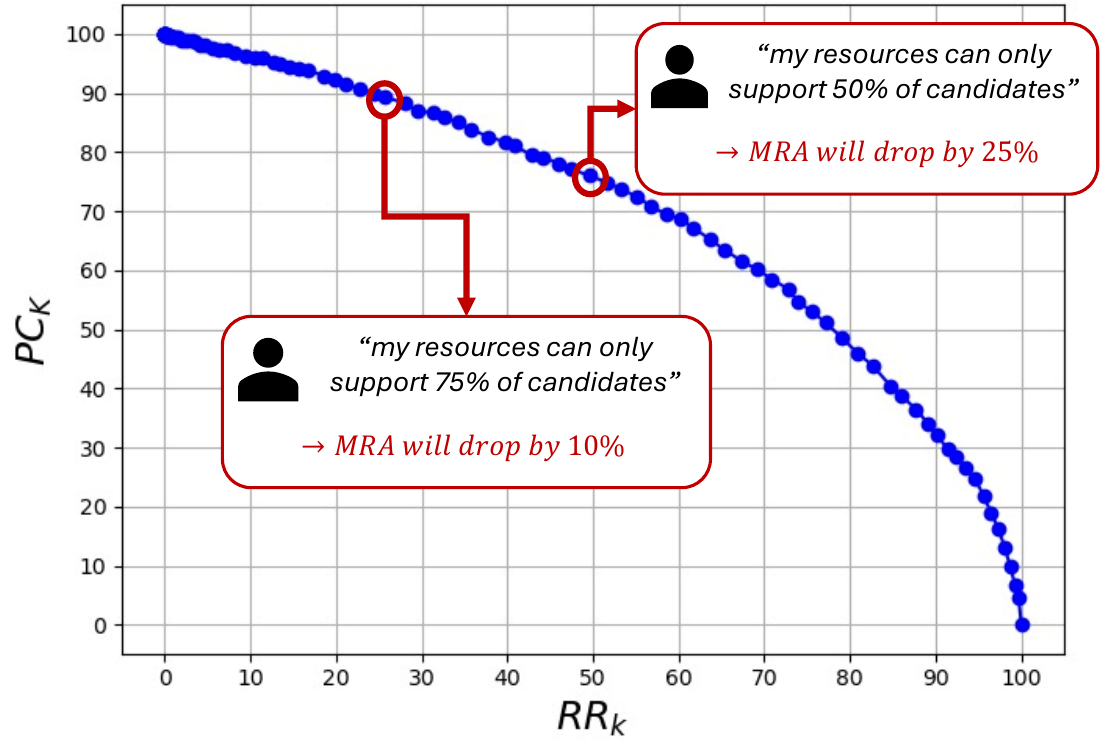}
    \caption{$PC_k$ vs. $RR_k$ on the PDM dataset, showing how threshold similarity-based pruning affects recall.}
    \label{fig:pc_k_vs_rr_k}
\end{figure}

\revGeneral{BKAFI has an optional pruning mechanism (see Section~\ref{subsection:blocking}) to accommodate resource constraints requiring smaller candidate sets. In this context, we use the following two metrics. The \emph{relative reduction ratio} ($RR_k$) measures the proportion of comparisons saved relative to the original blocking size: $RR_k = 1 - \frac{|C|}{k \cdot |D^C|}$. The \emph{relative pair completeness} ($PC_k$) captures the pair completeness after pruning, normalized by the unpruned value: $PC_k = \frac{\text{PC}(C)}{\text{PC}(k)}$, where $\text{PC}(C)$ and $\text{PC}(k)$ denote the number of true matches after and before pruning, respectively. These metrics help in estimating the MRA (recall in this case) loss with respect to constraints stated by disaster management experts, \emph{e.g.,} being able to handle only 75\% of the original candidate set size.}

\revGeneral{Figure~\ref{fig:pc_k_vs_rr_k} illustrates this tradeoff for our case study (the PDM dataset). Each point corresponds to a different similarity threshold, tracing the curve of achieved $PC_k$ as a function of $RR_k$. Minimal pruning ({\em i.e.}, low $RR_k$) retains nearly all true matches, resulting in near-perfect relative recall. Based on the figure, we can observe that if an expert wants to remove 25\% of the candidate pairs due to different constraints, the MRA (recall) drops by only $\sim$10\%. In scenarios that demand retaining only half of the candidates, the MRA drops by $\sim$25\%.} 

%% file: matching_experiments.tex
\subsection{Matching Evaluation}
\label{sec:matching_evaluation}
In this section, we evaluate the matching component of \model. 
We begin by comparing its performance with baseline methods (Section~\ref{subsection:exp_matching_baselines_experiment}). Next, we assess different machine learning models used as the backbone of \modelspace (Section~\ref{subsection:exp_matching_backbone_model_comparison}). \revA{Finally, we investigate performance over different contamination levels in Section~\ref{subsubsection:contamination_experiments}.}
\ifdefined \TechReport
Finally, we analyze the phenomenon of systematic discrepancy in Section~\ref{subsection:exp_discrepancy}.
\fi

\subsubsection{\modelspace VS. Baselines}
\label{subsection:exp_matching_baselines_experiment}
We compare \modelspace to baselines. 

\noindent \textbf{Performance:}
\label{subsubsection:exp_matching_baselines_experiment_performance}
Table~\ref{tab:matching_3dsager_vs_baselines} reports the matching performance of \modelspace and ViT-based baselines across three datasets. \modelspace consistently achieves the highest scores across all settings. On \emph{Hague\textsubscript{small}} and \emph{Hague\textsubscript{large}}, it reaches perfect precision with F1 scores of \revGeneral{$93.0$ and $92.8$}, respectively, outperforming both \emph{ViT-B/32} and \emph{ViT-L/14}. \revGeneral{On the \emph{PDM} dataset, \modelspace maintains strong performance ($F1 = 89.6$), while \emph{ViT-B/32} degrades significantly ($F1 = 63.9$). Results for \emph{ViT-L/14} on this dataset are omitted, as the model failed to converge, likely due to insufficient signal in the polygon mesh input and sensitivity to domain-specific variance.}

\begin{table}[t]
\centering
\caption{\revGeneral{Matching performance across datasets.}}
\scalebox{0.95}{\begin{tabular}{lcccc}
\toprule
\textbf{Dataset} & \textbf{Model} & \textbf{Pr.} & \textbf{Re.} & \textbf{F1} \\
\midrule
\multirow{3}{*}{\revGeneral{Hague\textsubscript{small}}}
  & \model       & \textbf{100.0}  & \textbf{87.0}  & \textbf{93.0} \\
  & ViT-B/32     & 92.4  & 85.2  & 88.6 \\
  & ViT-L/14     & 94.9  & 80.4  & 87.0 \\
\midrule
\multirow{3}{*}{\revGeneral{Hague\textsubscript{large}}}
  & \model       & \textbf{100.0}  & \textbf{86.6}  & \textbf{92.8} \\
  & ViT-B/32     & 88.4  & 83.8  & 86.1 \\
  & ViT-L/14     & 87.6  & 83.2  & 85.3 \\
\midrule
    \multirow{2}{*}{\revGeneral{PDM}} 
  & \model       & \textbf{94.2}  & \textbf{85.5}  & \textbf{89.6} \\
  & ViT-B/32     & 66.6  & 61.5  & 63.9 \\
  & ViT-L/14     & -  & -  & - \\
\bottomrule
\end{tabular}}
\label{tab:matching_3dsager_vs_baselines}
\end{table}

\begin{table}[b]
\centering
\caption{Runtime and space complexity on \emph{Hague\textsubscript{large}}.}
\scalebox{0.9}{\begin{tabular}{lcccccc}
\toprule
\textbf{Model} & \textbf{Vector} & \textbf{Train} & \textbf{Inf.} & \textbf{Model} & \textbf{\# Params /} \\
 & \textbf{Gen. (s)} & \textbf{(s)} & \textbf{(s)} & \textbf{(MB)} & \textbf{Nodes}  \\
& Train / Test &  &  &  &\\
\midrule
\textbf{3dSAGER} & \revGeneral{446.5 /} & \revGeneral{511.5} & \revGeneral{0.55} & 30.88 & 438,746 \\
\textbf{(Bagging)} & \revGeneral{270.7} &  &  & &  \\
\textbf{ViT-B/32} &  -- & \revGeneral{14,421} & \revGeneral{529} & 343.5 & 87,915,009 \\
\textbf{ViT-L/14} & -- & \revGeneral{53,055} & \revGeneral{2,402} & 1,188 & 304,064,769 \\
\bottomrule
\end{tabular}}
\label{tab:matching_complexity_comparison}
\end{table}

\noindent \textbf{Complexity:}
\label{subsubsection:exp_matching_baselines_experiment_complexity}
Table~\ref{tab:matching_complexity_comparison} compares the runtime (in seconds) and model complexity of \modelspace (using the best-performing \emph{Bagging} model) against ViT-based baselines. Experiments were conducted on the \emph{Hague\textsubscript{large}} dataset using a fixed random seed, with \revGeneral{60,393} training pairs and \revGeneral{52,600} test pairs. For \modelspace, we report the time required to generate property vectors for both splits. This step is unique to our pipeline and thus omitted for the baselines. Feature vector construction is computationally negligible and not reported separately. The training time for \modelspace includes grid search over a predefined hyperparameter space, meaning the actual per-configuration training time is substantially lower.

For \model, Evaluation is extremely fast, taking only a fraction of a second for all test pairs. The learned model has a compact footprint with around 439K nodes and a file size of just 30.88MB.
For the ViT-based models, we report the cumulative training time across all epochs. In our setup, embeddings are recomputed in each epoch (8 for ViT-B/32 and 5 for ViT-L/14), though in practice these can be cached to reduce overhead. Inference time refers to embedding generation and similarity scoring. ViT-B/32 and ViT-L/14 are significantly larger in size, with 88M and 304M parameters respectively. We do not account for the image rendering stage required to extract mesh images from CityJSON files for these baselines, as it is performed once and remains constant across runs.

The results highlight the efficiency advantage of \model. Training and inference are at least an order of magnitude faster than ViT-based models, with model size over $10\times$ smaller than \emph{ViT-B/32} and $40\times$ smaller than \emph{ViT-L/14}. Property vector generation, while non-negligible, is a one-time cost that scales linearly with the number of objects, making it suitable for large-scale use. In contrast, \emph{ViT} models require expensive fine-tuning and retain high inference overhead even with cached embeddings. While \modelspace consistently outperforms ViT-based baselines in matching accuracy, its key strength lies in its lightweight design, combining fast execution with a compact footprint. Together with the efficient BKAFI strategy (Section~\ref{subsubsection:execution_time_experiments}), these qualities make \modelspace a scalable and practical solution for real-world geospatial ER at scale.

\subsubsection{Backbone Model Comparison}
\label{subsection:exp_matching_backbone_model_comparison}
Table~\ref{tab:matching_backbone_models_comparison} compares the performance of various machine learning models used within \modelspace for \emph{Hague\textsubscript{small}} and \emph{PDM}. On \emph{Hague\textsubscript{small}}, \emph{Bagging} achieves the highest F1 score (\revGeneral{$93.0$}), with perfect precision, while \emph{XGBoost} and \emph{Gradient Boosting} also perform strongly, with F1 scores above \revGeneral{$89.5$}. \revGeneral{On the more challenging, real-world, \emph{PDM} dataset, \emph{XGBoost} yields the best overall performance ($F1 = 89.6$), followed by \emph{Gradient Boosting} and \emph{Bagging}. \emph{MLP} underperforms on both datasets, particularly in terms of recall, suggesting that neural models are less suited to the low-dimensional, structured feature space of our setting.} Overall, these results highlight the effectiveness and robustness of ensemble tree-based models for \fulltask.

\begin{table}[t]
\centering
\caption{Matching performance of different ML models in \modelspace across datasets.}
\begin{tabular}{lccc|ccc}
\toprule
\multirow{2}{*}{\textbf{Model}} & \multicolumn{3}{c|}{\textbf{Hague\textsubscript{small}}} & \multicolumn{3}{c}{\revGeneral{\textbf{PDM}}} \\
 & P & R & F1 & P & R & F1 \\
\midrule
Rand. Forest       & 89.5 & 87.8 & 88.7 & \textbf{95.9} & 80.0 & 87.2 \\
AdaBoost           & 85.9 & 85.7 & 85.8 & 94.2 & 80.3 & 86.7 \\
Gr. Boosting       & 89.8 & \textbf{89.3} & 89.5 & 94.7 & 84.1 & 89.1 \\
Bagging            & \textbf{100.0} & 87.0 & \textbf{93.0} & 95.6 & 82.4 & 88.5 \\
XGBoost            & 93.6 & 91.2 & 92.4 & 94.2 & \textbf{85.5} & \textbf{89.6} \\
MLP                & 62.6 & 68.2 & 65.2 & 93.4 & 56.4 & 70.0 \\
\bottomrule
\end{tabular}
\label{tab:matching_backbone_models_comparison}
\end{table}

\subsubsection{\modelspace Over Contaminated datasets}
\label{subsubsection:contamination_experiments}
\revA{To evaluate the robustness of \modelspace in settings where the dataset is contaminated, we simulate controlled contamination by introducing cross-source reference swaps into  the \emph{Hague\textsubscript{small}} dataset. This is done by swapping a percentage of items between the candidate and index sets, in accordance with a predefined contamination level, varying from 0.05 to 0.5 in increments of 0.05. This setup weakens the systematic discrepancy assumption, as candidate-index alignment becomes noisier, a condition that mirrors post-disaster urban reconstruction scenarios where data sources may be heterogeneous or partially overlapping. We evaluate the F1 score for three classifiers (Random Forest, Bagging, and XGBoost) as the backbone matcher.
}

\begin{figure}[b]
    \centering
    \includegraphics[width=0.998\columnwidth]{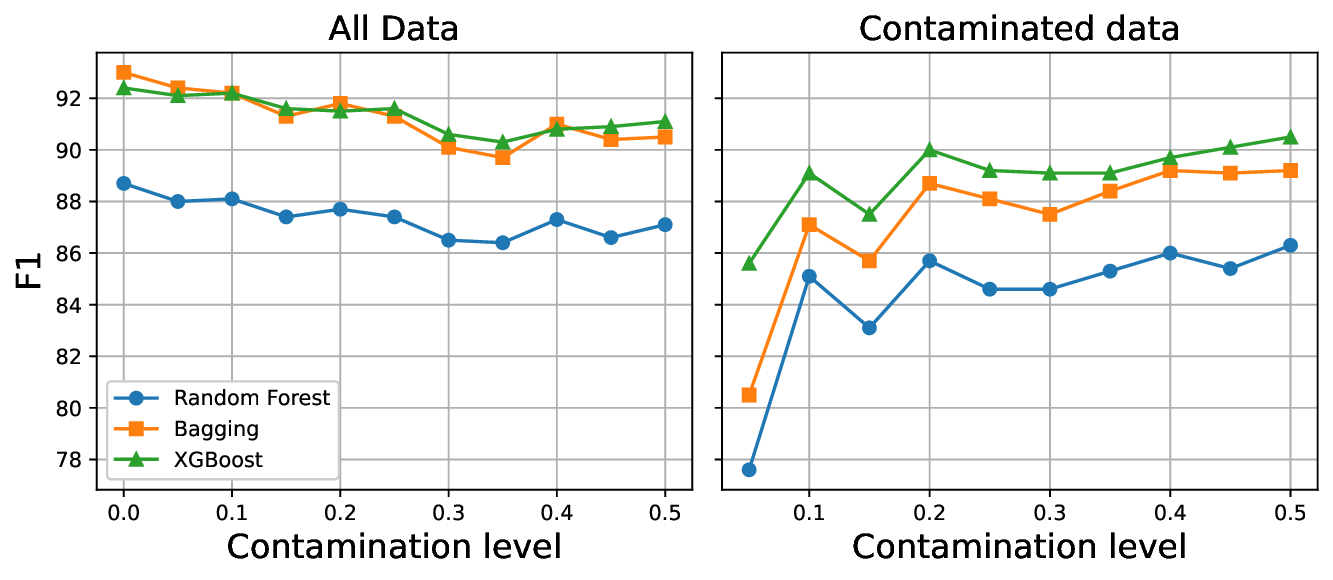}
    \caption{\revA{Robustness of \modelspace to increasing contamination levels on the \emph{Hague\textsubscript{small}} dataset. Left: overall F1. Right: F1 over contaminated samples only.}}
    \label{fig:f1_vs_contamination_combined}
\end{figure}

\revA{
Figure~\ref{fig:f1_vs_contamination_combined} (left) provides the overall F1 score across all test samples as a function of contamination level. Increasing contamination degrades overall performance due to the difficulty of learning consistent matching patterns under noisy alignment. Still, \modelspace exhibits robust behavior, with all backbone models maintaining a relatively slight performance drop (\emph{e.g.}, XGBoost decreases by only 1.3, from 92.4 at 0\% contamination to 91.1 at 50\%). }

\revA{
When isolating performance over the contaminated portion of the test set (Figure~\ref{fig:f1_vs_contamination_combined}, right), we observe the opposite trend: F1 improves as contamination increases. This suggests that as the model is exposed to more contaminated examples during training, it becomes better equipped to identify such cases at test time. For all backbone models, \modelspace stabilizes after a relatively small amount of contamination, indicating its capacity to adapt quickly to noisy conditions. }

\subsubsection{Dirty-Clean Setting: Mixed-Source Transferability Analysis}
\label{subsubsection:dirty_clean_transferability}
To further challenge our framework, we depart from the clean-clean assumption by introducing a mixed-source setting in which the candidate and index sets are no longer strictly disjoint. Specifically, we select objects from the candidate set that have known matches in the index set and move their corresponding matches into the candidate set. This process is repeated with increased contamination level ranging from 0.05 to 0.5. By the, we introduce ambiguity into the matching structure, as some entities now appear in the same source (candidate set). Beyond simulating noisy alignment, this setup also serves to examine the \emph{transferability} of the trained matcher. Here, the pipeline is executed entirely within the contaminated candidate set, and the resulting matcher is then applied to the cross-source test set. This allows us to assess whether knowledge learned in a noisy, within-source scenario can be effectively transferred to the standard cross-source matching setting.

\begin{figure}[t]
    \centering
    \includegraphics[width=0.92961\columnwidth]{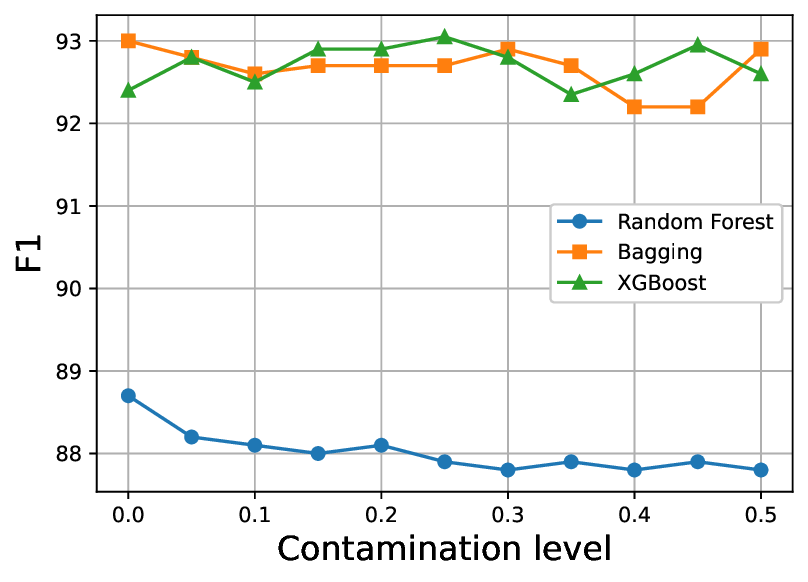}
    \caption{Performance (F1 score) of Random Forest, Bagging, and XGBoost backbone matchers in the dirty-clean setting across different contamination levels.}
    \label{fig:f1_vs_dirty_cont}
\end{figure}

Figure~\ref{fig:f1_vs_dirty_cont} summarizes the results. All three backbone matchers experience only a slight drop in F1 when moving from the clean-clean setting (contamination level 0) to the highest contamination level (0.5), indicating that the transfer from the noisy training environment to the clean test setting is largely successful. The Random Forest matcher shows a gradual decline from about 88.7 to 87.8, whereas Bagging and XGBoost maintain are relatively consistent, with no clear sensitivity to the contamination level. These findings confirm our initial expectation that contamination in training leads to only minor degradation in clean-test performance.


\ifdefined\TechReport
\subsubsection{Systematic Discrepancy Discussion}
\label{subsection:exp_discrepancy}
Recall that $(\epsilon, \delta)$-systematic discrepancy (Definition~\ref{definition:systematic_discrepancy}) characterizes how consistently a geometric property exhibits a stable ratio $r_g$ across matching pairs. Here, $\epsilon$ defines the tolerance around $r_g$, and $\delta$ denotes the fraction of matches exceeding this tolerance. We next qualitatively explore this behavior by analyzing how $\delta$ varies with $\epsilon$ for several top-ranked properties (based on feature importance scores from a single run on \emph{Hague\textsubscript{large}} using the best-performing matcher).

\begin{figure}[t]
    \centering
    \includegraphics[width=0.908\linewidth]{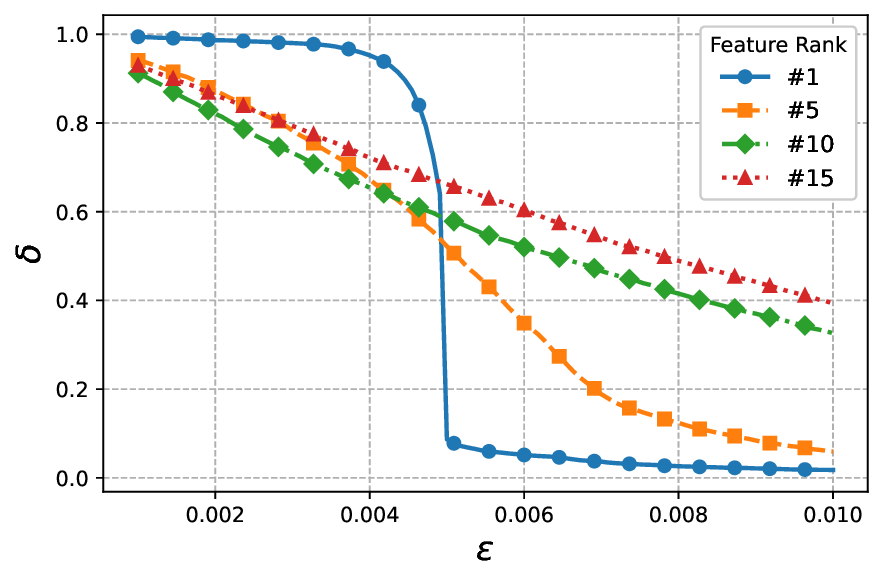}
    \caption{$(\epsilon, \delta)$ curves for selected geometric properties (indexed by importance rank). Lower curves reflect stronger systematic discrepancy.}
    \label{fig:eps_delta_curves}
\end{figure}

Figure~\ref{fig:eps_delta_curves} shows $(\epsilon, \delta)$ plots for four representative geometric properties, ranked by their feature importance score (first, fifth, etc.). Each curve reflects the trade-off between tolerance ($\epsilon$) and the proportion of matching pairs that violate the discrepancy condition ($\delta$). A sharper drop in $\delta$ (\emph{e.g.,} $\#1$ and $\#5$) indicates properties with strong systematic patterns, meaning that most matching pairs follow a consistent ratio, even under tight $\epsilon$ bounds. In contrast, smoother curves (e.g., $\#10$ and $\#15$) reveal properties with higher dispersion and weaker consistency across sources. These trends support the use of systematic discrepancy correction for select properties, especially those with low $\delta$ under small $\epsilon$ values.

To further elucidate the meaning of ratio distribution, we present Figure~\ref{fig:ratio_dist_comparison} that compares the ratio distributions of the two top-ranked geometric properties -- \texttt{bounding\_box\_width} and \texttt{perimeter} -- between matching and non-matching pairs. As seen on the left, the ratios for matching pairs are highly concentrated around 1.0, indicating strong consistency. Conversely, the distributions for non-matching pairs (right) are notably broader and more irregular, lacking the sharp central peak. This highlights a key motivation behind our geometric featurization: for certain properties, matching pairs exhibit distinctive and predictable proportional patterns that are absent in non-matching pairs. This disparity enables effective discrimination by the downstream model.

\begin{figure}[t]
    \centering
    \includegraphics[width=0.99\columnwidth]{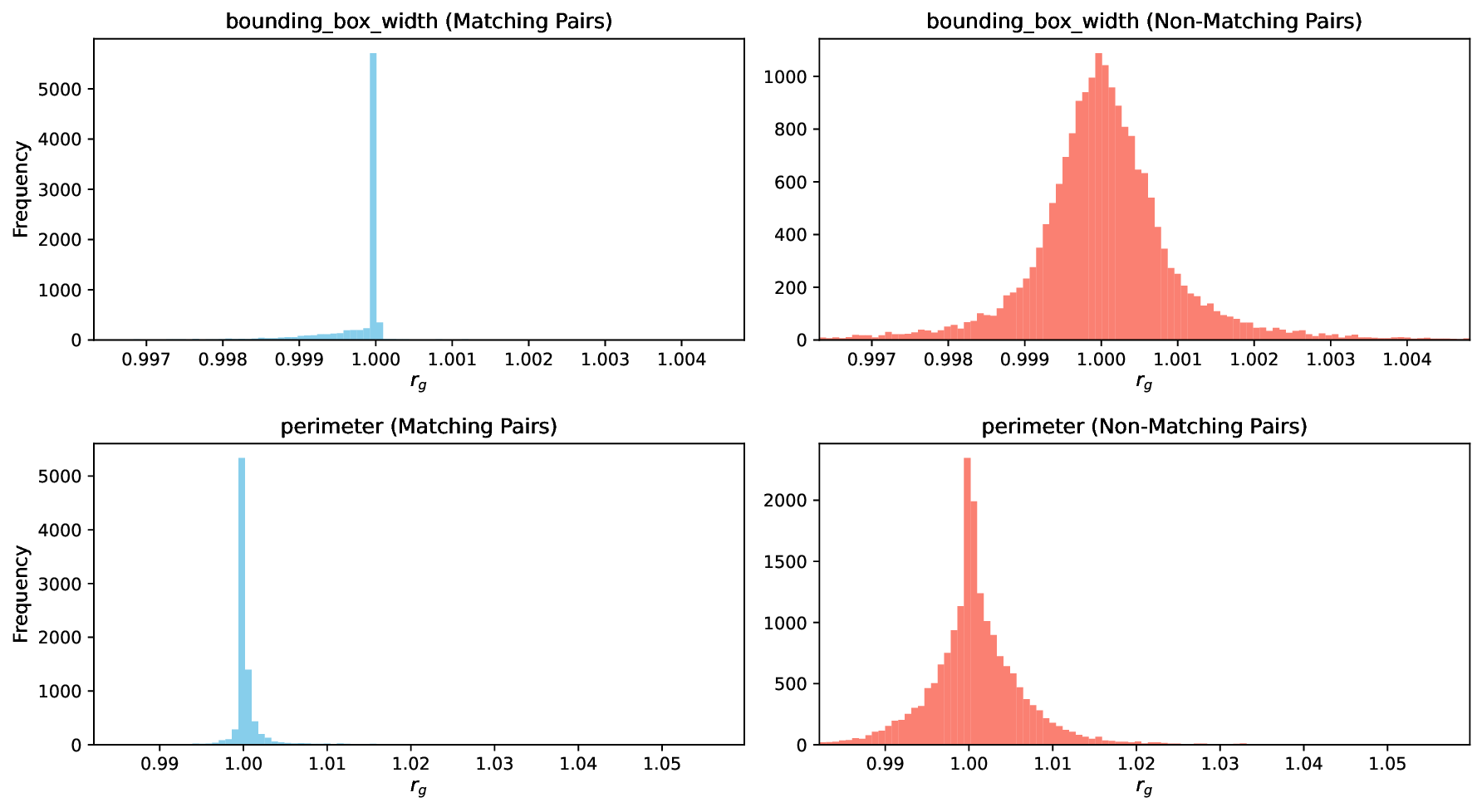}
    \caption{Ratio distributions for top features among matching vs. non-matching pairs.}
    \label{fig:ratio_dist_comparison}
\end{figure}
\fi

%% file: related_work.tex
\section{Related Work}
\label{sec:related_work}
ER is among the most 
widely-studied data integration tasks, aimed at identifying correspondences between data records that refer to the same real-world entity~\cite{cohen2002learning, christen2012data, winkler2002methods, getoor2012entity}. 
ER typically consists of two main steps: blocking and matching. The former reduces the number of candidate comparisons by grouping likely matches based on predefined attributes or heuristics~\cite{papadakis2020blocking, li2020survey, papadakis2011eliminating}. The latter determines whether each candidate pair is indeed a true match. Solutions for ER have evolved over the years, starting from relying on string similarity between textual elements~\cite{levenshtein1966binary,jaro1989advances,lin1998information,jaro1995probabilistic} and probabilistic approaches~\cite{fellegi1969theory}, and followed by rule-based methods~\cite{singla2006entity,singh2017synthesizing}. 
As the complexity of data sources increased and the volume of records grew, ER methods adopted machine learning techniques, offering greater flexibility in modeling complex relationships in data~\cite{bilenko2003adaptive,konda2016magellan}. Recent approaches increasingly rely on deep learning~\cite{mudgal2018deep, joty2018distributed}, especially leveraging pretrained language models for textual data~\cite{li2020deep, li2021deep, peeters2023using, peeters2023entity, wang2022promptem}. \revA{We now provide a more detailed discussion on Geospatial ER, 3D Data Featurization and, given the novelty of BKAFI, blocking techniques.}

\revA{
\noindent \textbf{Geospatial ER}: 
Geospatial ER and its extensions exemplify use cases that extend beyond tabular or textual data sources~\cite{sehgal2006entity, kang2007geoddupe}. Some existing works focus on matching objects based on geographic information, (\emph{i.e.,} coordinates), obtained from sensors~\cite{tabarro2017webgis, balley2004modelling}. Another line of works suggests to incorporate signals such as textual location descriptions. Balsebre \emph{et al.}~\cite{balsebre2022geospatial} presented Geo-ER, in which graph neural networks generate a unified entity representations based on object textual attributes along with their location. Barret \emph{et al.}~\cite{barret2019spatial} introduced GeoAlign, a system that allows users to customize the desired similarity measures and attributes to be used for the matching decision. Closest to our work is that of Shah \emph{et al.}~\cite{shah2021gem} that use polygons in 2D space, rather than a single coordinate, to model an object. They describe spatial relationships between entities based on distance measures, as well as spatial proximity-based features, such as overlaps and containment, enabling more nuanced matching decisions.
All existing works operate in 2D space and rely on flat spatial representations, such as centroids, bounding boxes, or polygon overlays, along with metadata like textual descriptions. While these representations suffice for relatively simple geospatial objects, they fall short in capturing the structural complexity of real-world 3D environments. We use 3D polygon mesh representations, encoding detailed geometric and topological information about spatial entities. These richer representations enable the extraction of intrinsic geometric properties that are inaccessible in 2D. The rich representation motivates a shift in matching paradigm: from location-based or textual similarity to similarity in a geometric property space derived directly from raw 3D representation.
}

\revA{
\noindent \textbf{3D Data Featurization}:
Several works explore the use of geometric features over 3D spatial building data, primarily in the context of urban modeling. For example, Labetski \emph{et al.}~\cite{labetski20233d} review a wide range of applications in which 3D building metrics are used for tasks like energy consumption estimation and urban morphology analysis, motivating the use of 3D metrics in reducing ambiguities and supporting city complexity parametrization. Jaljolie \emph{et al.}~\cite{jaljolie2018spatial} propose a spatial data structure for 3D land management, introducing hierarchical 3D primitives and topological reasoning for volumetric parcels. They also develop a topological-based algorithm to classify spatial relationships among complex 3D geometries~\cite{jaljolie2021topological}. While these works demonstrate the utility of geometric and topological 3D features in land administration and urban planning, they do not address ER or operate directly on raw polygon mesh inputs. Our work fills this gap by generating geometric featurization from raw 3D, which dictates our novel problem formulation. This connection between raw 3D data and derived feature space is key to the design of  BKAFI, which leverages this structure to enable efficient and interpretable candidate generation.
}

\revA{
\noindent \textbf{Blocking Techniques}:
Traditional blocking methods rely on cues from structured data, {\em e.g.}, q-grams~\cite{kenig2013mfiblocks, papadakis2015schema, paulsen2023sparkly}. In our case, such signals are absent, as inputs are unstructured polygon meshes. Modern approaches increasingly rely on embedding-based representations to capture semantic similarity between entities~\cite{thirumuruganathan2021deep, brinkmann2024sc, zhang2020autoblock}. These embeddings project entities into a continuous vector space, where geometric distance serves as a proxy for similarity, a notion that has also been adopted in related ER sub-tasks, such as active learning~\cite{jain2021deep, genossar2023battleship}. This vector-based reasoning is also central to unsupervised pipelines such as ZeroER~\cite{wu2020zeroer}, which construct similarity-based feature vectors over entity pairs to enable learning without labeled data.
Existing works propose using variations of contrastive learning to train a blocker~\cite{jain2021deep, brinkmann2024sc, thirumuruganathan2021deep}. These approaches operate on tabular data with textual attributes and typically employ attention-based encoders to generate record embeddings. The training objective is to pull embeddings of matching pairs closer in the vector space while pushing apart non-matching pairs. In contrast, Bilenko \emph{et al.}~\cite{bilenko2003adaptive} introduce adaptive blocking functions based on data-derived predicates.
In geospatial ER, blocking methods often assume that references to the same entity lie near each other in the coordinate space, leading to pointwise similarity techniques~\cite{balsebre2022geospatial, kang2007geoddupe, khodizadeh2021novel}. However, in the absence of coordinate systems or in the presence of spatial distortions, 
performance significantly deteriorates, motivating the need for alternative strategies.
BKAFI departs from prior trainable blocking methods in several key ways. First, while existing approaches learn blocking functions from raw textual attributes or embedding distances ({\em e.g.}, using contrastive learning), BKAFI leverages a trained matcher to identify the most informative geometric properties. Blocking keys are selected based on feature importance scores, focusing on a meaningful, learned subspace. Second, while traditional geospatial approaches assume spatial proximity based on flat representations such as object centroids, BKAFI operates entirely in the geometric feature space derived from 3D polygon meshes. This design allows BKAFI to remain effective in settings where coordinate systems are unreliable or unavailable, while also ensuring both efficiency and interpretability.
}

%% file: conclusions.tex
\section{Conclusions}
\label{sec:conclusions}
In this work, we introduce \model, an end-to-end pipeline for geospatial entity resolution over 3D objects. We levarage a novel featurization process that captures the geometric characteristics of 3D polygon meshes and serves as the foundation for both the matching model and our novel blocking technique, BKAFI. To support evaluation and encourage future research, we curated a new benchmark for geospatial ER over 3D objects. Empirical results on the new benchmark, alongside a proprietary dataset, demonstrate the effectiveness and efficiency of our approach in addressing the challenges of both components of the \taskspace pipelines

In future work, we plan to experiment with datasets exhibiting varying levels of detail, and expanding the design of the pairwise feature vectors to capture a broader range of spatial characteristics. In particular, we aim to explore reference semantics using a graph representation learning framework capable of modeling richer structural information embedded in 3D models, such as windows and doors within walls and roofs. Another future work direction will examine properties and guarantees of BKAFI, with the aim of extending its applicability to other vector-search applications.

%% file: acknowledgements.tex
\section*{Acknowledgments}
We acknowledge the use of OpenAI's ChatGPT for assistance in the preparation of this manuscript. Specifically, the tool was used for rephrasing parts of the text, converting raw experimental results into \LaTeX\ tables, and generating Python code for plotting figures.
This work was supported in part by Research \& Development grant 2033008. Shraga's work was supported in part by NSF under award number IIS-2325632.

%% file: main.bbl

\begin{thebibliography}{82}


\ifx \showCODEN    \undefined \def \showCODEN     #1{\unskip}     \fi
\ifx \showISBNx    \undefined \def \showISBNx     #1{\unskip}     \fi
\ifx \showISBNxiii \undefined \def \showISBNxiii  #1{\unskip}     \fi
\ifx \showISSN     \undefined \def \showISSN      #1{\unskip}     \fi
\ifx \showLCCN     \undefined \def \showLCCN      #1{\unskip}     \fi
\ifx \shownote     \undefined \def \shownote      #1{#1}          \fi
\ifx \showarticletitle \undefined \def \showarticletitle #1{#1}   \fi
\ifx \showURL      \undefined \def \showURL       {\relax}        \fi
\providecommand\bibfield[2]{#2}
\providecommand\bibinfo[2]{#2}
\providecommand\natexlab[1]{#1}
\providecommand\showeprint[2][]{arXiv:#2}

\bibitem[Adams et~al\mbox{.}(2008)]%
        {Adams2008TheAP}
\bibfield{author}{\bibinfo{person}{Niall~M. Adams}, \bibinfo{person}{Martin
  Field}, \bibinfo{person}{Erol Gelenbe}, \bibinfo{person}{David~J. Hand},
  \bibinfo{person}{Nicholas~R. Jennings}, \bibinfo{person}{David~S. Leslie},
  \bibinfo{person}{David Nicholson}, \bibinfo{person}{Sarvapali~D. Ramchurn},
  \bibinfo{person}{Stephen~J. Roberts}, {and} \bibinfo{person}{Alex Rogers}.}
  \bibinfo{year}{2008}\natexlab{}.
\newblock \showarticletitle{The ALADDIN project : intelligent agents for
  disaster management}.
\newblock
\urldef\tempurl%
\url{https://api.semanticscholar.org/CorpusID:18836574}
\showURL{%
\tempurl}


\bibitem[Balley et~al\mbox{.}(2004)]%
        {balley2004modelling}
\bibfield{author}{\bibinfo{person}{Sandrine Balley}, \bibinfo{person}{Christine
  Parent}, {and} \bibinfo{person}{Stefano Spaccapietra}.}
  \bibinfo{year}{2004}\natexlab{}.
\newblock \showarticletitle{Modelling geographic data with multiple
  representations}.
\newblock \bibinfo{journal}{\emph{International Journal of Geographical
  Information Science}} \bibinfo{volume}{18}, \bibinfo{number}{4}
  (\bibinfo{year}{2004}), \bibinfo{pages}{327--352}.
\newblock


\bibitem[Balouek-Thomert et~al\mbox{.}(2022)]%
        {balouek2022towards}
\bibfield{author}{\bibinfo{person}{Daniel Balouek-Thomert},
  \bibinfo{person}{Eddy Caron}, \bibinfo{person}{Laurent Lefevre}, {and}
  \bibinfo{person}{Manish Parashar}.} \bibinfo{year}{2022}\natexlab{}.
\newblock \showarticletitle{Towards a methodology for building dynamic urgent
  applications on continuum computing platforms}. In
  \bibinfo{booktitle}{\emph{2022 First Combined International Workshop on
  Interactive Urgent Supercomputing (CIW-IUS)}}. IEEE, \bibinfo{pages}{1--6}.
\newblock


\bibitem[Balsebre et~al\mbox{.}(2022)]%
        {balsebre2022geospatial}
\bibfield{author}{\bibinfo{person}{Pasquale Balsebre}, \bibinfo{person}{Dezhong
  Yao}, \bibinfo{person}{Gao Cong}, {and} \bibinfo{person}{Zhen Hai}.}
  \bibinfo{year}{2022}\natexlab{}.
\newblock \showarticletitle{Geospatial entity resolution}. In
  \bibinfo{booktitle}{\emph{Proceedings of the ACM Web Conference 2022}}.
  \bibinfo{pages}{3061--3070}.
\newblock


\bibitem[Barret et~al\mbox{.}(2019)]%
        {barret2019spatial}
\bibfield{author}{\bibinfo{person}{Nelly Barret}, \bibinfo{person}{Fabien
  Duchateau}, \bibinfo{person}{Franck Favetta}, {and} \bibinfo{person}{Ludovic
  Moncla}.} \bibinfo{year}{2019}\natexlab{}.
\newblock \showarticletitle{Spatial entity matching with geoalign (demo
  paper)}. In \bibinfo{booktitle}{\emph{Proceedings of the 27th ACM SIGSPATIAL
  International Conference on Advances in Geographic Information Systems}}.
  \bibinfo{pages}{580--583}.
\newblock


\bibitem[Bentley(1975)]%
        {bentley1975multidimensional}
\bibfield{author}{\bibinfo{person}{Jon~Louis Bentley}.}
  \bibinfo{year}{1975}\natexlab{}.
\newblock \showarticletitle{Multidimensional binary search trees used for
  associative searching}.
\newblock \bibinfo{journal}{\emph{Commun. ACM}} \bibinfo{volume}{18},
  \bibinfo{number}{9} (\bibinfo{year}{1975}), \bibinfo{pages}{509--517}.
\newblock


\bibitem[Bilenko and Mooney(2003)]%
        {bilenko2003adaptive}
\bibfield{author}{\bibinfo{person}{Mikhail Bilenko} {and}
  \bibinfo{person}{Raymond~J Mooney}.} \bibinfo{year}{2003}\natexlab{}.
\newblock \showarticletitle{Adaptive duplicate detection using learnable string
  similarity measures}. In \bibinfo{booktitle}{\emph{Proceedings of the ninth
  ACM SIGKDD international conference on Knowledge discovery and data mining}}.
  \bibinfo{pages}{39--48}.
\newblock


\bibitem[Bouzas et~al\mbox{.}(2020)]%
        {bouzas2020structure}
\bibfield{author}{\bibinfo{person}{Vasileios Bouzas}, \bibinfo{person}{Hugo
  Ledoux}, {and} \bibinfo{person}{Liangliang Nan}.}
  \bibinfo{year}{2020}\natexlab{}.
\newblock \showarticletitle{Structure-aware building mesh polygonization}.
\newblock \bibinfo{journal}{\emph{ISPRS Journal of Photogrammetry and Remote
  Sensing}}  \bibinfo{volume}{167} (\bibinfo{year}{2020}),
  \bibinfo{pages}{432--442}.
\newblock


\bibitem[Breiman(1996)]%
        {breiman1996bagging}
\bibfield{author}{\bibinfo{person}{Leo Breiman}.}
  \bibinfo{year}{1996}\natexlab{}.
\newblock \showarticletitle{Bagging predictors}.
\newblock \bibinfo{journal}{\emph{Machine learning}}  \bibinfo{volume}{24}
  (\bibinfo{year}{1996}), \bibinfo{pages}{123--140}.
\newblock


\bibitem[Breiman(2001)]%
        {breiman2001random}
\bibfield{author}{\bibinfo{person}{Leo Breiman}.}
  \bibinfo{year}{2001}\natexlab{}.
\newblock \showarticletitle{Random forests}.
\newblock \bibinfo{journal}{\emph{Machine learning}}  \bibinfo{volume}{45}
  (\bibinfo{year}{2001}), \bibinfo{pages}{5--32}.
\newblock


\bibitem[Brinkmann et~al\mbox{.}(2024)]%
        {brinkmann2024sc}
\bibfield{author}{\bibinfo{person}{Alexander Brinkmann}, \bibinfo{person}{Roee
  Shraga}, {and} \bibinfo{person}{Christina Bizer}.}
  \bibinfo{year}{2024}\natexlab{}.
\newblock \showarticletitle{Sc-block: Supervised contrastive blocking within
  entity resolution pipelines}. In \bibinfo{booktitle}{\emph{European Semantic
  Web Conference}}. Springer, \bibinfo{pages}{121--142}.
\newblock


\bibitem[Changyong et~al\mbox{.}(2014)]%
        {changyong2014log}
\bibfield{author}{\bibinfo{person}{FENG Changyong}, \bibinfo{person}{WANG
  Hongyue}, \bibinfo{person}{LU Naiji}, \bibinfo{person}{CHEN Tian},
  \bibinfo{person}{HE Hua}, \bibinfo{person}{LU Ying}, {and}
  \bibinfo{person}{M~TU Xin}.} \bibinfo{year}{2014}\natexlab{}.
\newblock \showarticletitle{Log-transformation and its implications for data
  analysis}.
\newblock \bibinfo{journal}{\emph{Shanghai archives of psychiatry}}
  \bibinfo{volume}{26}, \bibinfo{number}{2} (\bibinfo{year}{2014}),
  \bibinfo{pages}{105}.
\newblock


\bibitem[Chen and Guestrin(2016)]%
        {chen2016xgboost}
\bibfield{author}{\bibinfo{person}{Tianqi Chen} {and} \bibinfo{person}{Carlos
  Guestrin}.} \bibinfo{year}{2016}\natexlab{}.
\newblock \showarticletitle{Xgboost: A scalable tree boosting system}. In
  \bibinfo{booktitle}{\emph{Proceedings of the 22nd acm sigkdd international
  conference on knowledge discovery and data mining}}.
  \bibinfo{pages}{785--794}.
\newblock


\bibitem[Chiabrando et~al\mbox{.}(2019)]%
        {chiabrando2019uav}
\bibfield{author}{\bibinfo{person}{F Chiabrando}, \bibinfo{person}{F
  Giulio~Tonolo}, {and} \bibinfo{person}{Andrea Lingua}.}
  \bibinfo{year}{2019}\natexlab{}.
\newblock \showarticletitle{Uav direct georeferencing approach in an emergency
  mapping context. the 2016 central Italy earthquake case study}.
\newblock \bibinfo{journal}{\emph{The International Archives of the
  Photogrammetry, Remote Sensing and Spatial Information Sciences}}
  \bibinfo{volume}{42} (\bibinfo{year}{2019}), \bibinfo{pages}{247--253}.
\newblock


\bibitem[Christen and Christen(2012)]%
        {christen2012data}
\bibfield{author}{\bibinfo{person}{Peter Christen} {and} \bibinfo{person}{Peter
  Christen}.} \bibinfo{year}{2012}\natexlab{}.
\newblock \bibinfo{booktitle}{\emph{The data matching process}}.
\newblock \bibinfo{publisher}{Springer}.
\newblock


\bibitem[Cohen and Richman(2002)]%
        {cohen2002learning}
\bibfield{author}{\bibinfo{person}{William~W Cohen} {and}
  \bibinfo{person}{Jacob Richman}.} \bibinfo{year}{2002}\natexlab{}.
\newblock \showarticletitle{Learning to match and cluster large
  high-dimensional data sets for data integration}. In
  \bibinfo{booktitle}{\emph{Proceedings of the eighth ACM SIGKDD international
  conference on Knowledge discovery and data mining}}.
  \bibinfo{pages}{475--480}.
\newblock


\bibitem[Deng et~al\mbox{.}(2019)]%
        {deng2019point}
\bibfield{author}{\bibinfo{person}{Yue Deng}, \bibinfo{person}{An Luo},
  \bibinfo{person}{Jiping Liu}, {and} \bibinfo{person}{Yong Wang}.}
  \bibinfo{year}{2019}\natexlab{}.
\newblock \showarticletitle{Point of interest matching between different
  geospatial datasets}.
\newblock \bibinfo{journal}{\emph{ISPRS International Journal of
  Geo-Information}} \bibinfo{volume}{8}, \bibinfo{number}{10}
  (\bibinfo{year}{2019}), \bibinfo{pages}{435}.
\newblock


\bibitem[D{\"o}llner and Hagedorn(2007)]%
        {dollner2007integrating}
\bibfield{author}{\bibinfo{person}{J{\"u}rgen D{\"o}llner} {and}
  \bibinfo{person}{Benjamin Hagedorn}.} \bibinfo{year}{2007}\natexlab{}.
\newblock \showarticletitle{Integrating urban GIS, CAD, and BIM data by
  service-based virtual 3D city models}.
\newblock In \bibinfo{booktitle}{\emph{Urban and regional data management}}.
  \bibinfo{publisher}{CRC Press}, \bibinfo{pages}{157--170}.
\newblock


\bibitem[Dosovitskiy et~al\mbox{.}(2021)]%
        {dosovitskiy2020vit}
\bibfield{author}{\bibinfo{person}{Alexey Dosovitskiy}, \bibinfo{person}{Lucas
  Beyer}, \bibinfo{person}{Alexander Kolesnikov}, \bibinfo{person}{Dirk
  Weissenborn}, \bibinfo{person}{Xiaohua Zhai}, \bibinfo{person}{Thomas
  Unterthiner}, \bibinfo{person}{Mostafa Dehghani}, \bibinfo{person}{Matthias
  Minderer}, \bibinfo{person}{Georg Heigold}, \bibinfo{person}{Sylvain Gelly},
  \bibinfo{person}{Jakob Uszkoreit}, {and} \bibinfo{person}{Neil Houlsby}.}
  \bibinfo{year}{2021}\natexlab{}.
\newblock \showarticletitle{An Image is Worth 16x16 Words: Transformers for
  Image Recognition at Scale}.
\newblock \bibinfo{journal}{\emph{ICLR}} (\bibinfo{year}{2021}).
\newblock


\bibitem[Elmagarmid et~al\mbox{.}(2006)]%
        {elmagarmid2006duplicate}
\bibfield{author}{\bibinfo{person}{Ahmed~K Elmagarmid},
  \bibinfo{person}{Panagiotis~G Ipeirotis}, {and} \bibinfo{person}{Vassilios~S
  Verykios}.} \bibinfo{year}{2006}\natexlab{}.
\newblock \showarticletitle{Duplicate record detection: A survey}.
\newblock \bibinfo{journal}{\emph{IEEE Transactions on knowledge and data
  engineering}} \bibinfo{volume}{19}, \bibinfo{number}{1}
  (\bibinfo{year}{2006}), \bibinfo{pages}{1--16}.
\newblock


\bibitem[Erving et~al\mbox{.}(2009)]%
        {erving2009data}
\bibfield{author}{\bibinfo{person}{Anna Erving}, \bibinfo{person}{Petri
  R{\"o}nnholm}, {and} \bibinfo{person}{Milka Nuikka}.}
  \bibinfo{year}{2009}\natexlab{}.
\newblock \showarticletitle{Data integration from different sources to create
  3D virtual model}.
\newblock \bibinfo{journal}{\emph{3D-ARCH}}  \bibinfo{volume}{2009}
  (\bibinfo{year}{2009}), \bibinfo{pages}{3D}.
\newblock


\bibitem[Fellegi and Sunter(1969)]%
        {fellegi1969theory}
\bibfield{author}{\bibinfo{person}{Ivan~P Fellegi} {and}
  \bibinfo{person}{Alan~B Sunter}.} \bibinfo{year}{1969}\natexlab{}.
\newblock \showarticletitle{A theory for record linkage}.
\newblock \bibinfo{journal}{\emph{J. Amer. Statist. Assoc.}}
  \bibinfo{volume}{64}, \bibinfo{number}{328} (\bibinfo{year}{1969}),
  \bibinfo{pages}{1183--1210}.
\newblock


\bibitem[Freund et~al\mbox{.}(1996)]%
        {freund1996experiments}
\bibfield{author}{\bibinfo{person}{Yoav Freund}, \bibinfo{person}{Robert~E
  Schapire}, {et~al\mbox{.}}} \bibinfo{year}{1996}\natexlab{}.
\newblock \showarticletitle{Experiments with a new boosting algorithm}. In
  \bibinfo{booktitle}{\emph{icml}}, Vol.~\bibinfo{volume}{96}. Citeseer,
  \bibinfo{pages}{148--156}.
\newblock


\bibitem[Friedman(2001)]%
        {friedman2001greedy}
\bibfield{author}{\bibinfo{person}{Jerome~H Friedman}.}
  \bibinfo{year}{2001}\natexlab{}.
\newblock \showarticletitle{Greedy function approximation: a gradient boosting
  machine}.
\newblock \bibinfo{journal}{\emph{Annals of statistics}}
  (\bibinfo{year}{2001}), \bibinfo{pages}{1189--1232}.
\newblock


\bibitem[Genossar et~al\mbox{.}(2023)]%
        {genossar2023battleship}
\bibfield{author}{\bibinfo{person}{Bar Genossar}, \bibinfo{person}{Avigdor
  Gal}, {and} \bibinfo{person}{Roee Shraga}.} \bibinfo{year}{2023}\natexlab{}.
\newblock \showarticletitle{The battleship approach to the low resource entity
  matching problem}.
\newblock \bibinfo{journal}{\emph{Proceedings of the ACM on Management of
  Data}} \bibinfo{volume}{1}, \bibinfo{number}{4} (\bibinfo{year}{2023}),
  \bibinfo{pages}{1--25}.
\newblock


\bibitem[Getoor and Diehl(2005)]%
        {getoor2005link}
\bibfield{author}{\bibinfo{person}{Lise Getoor} {and}
  \bibinfo{person}{Christopher~P Diehl}.} \bibinfo{year}{2005}\natexlab{}.
\newblock \showarticletitle{Link mining: a survey}.
\newblock \bibinfo{journal}{\emph{Acm Sigkdd Explorations Newsletter}}
  \bibinfo{volume}{7}, \bibinfo{number}{2} (\bibinfo{year}{2005}),
  \bibinfo{pages}{3--12}.
\newblock


\bibitem[Getoor and Machanavajjhala(2012)]%
        {getoor2012entity}
\bibfield{author}{\bibinfo{person}{Lise Getoor} {and} \bibinfo{person}{Ashwin
  Machanavajjhala}.} \bibinfo{year}{2012}\natexlab{}.
\newblock \showarticletitle{Entity resolution: theory, practice \& open
  challenges}.
\newblock \bibinfo{journal}{\emph{Proceedings of the VLDB Endowment}}
  \bibinfo{volume}{5}, \bibinfo{number}{12} (\bibinfo{year}{2012}),
  \bibinfo{pages}{2018--2019}.
\newblock


\bibitem[Gr{\"o}ger and Pl{\"u}mer(2012)]%
        {groger2012citygml}
\bibfield{author}{\bibinfo{person}{Gerhard Gr{\"o}ger} {and}
  \bibinfo{person}{Lutz Pl{\"u}mer}.} \bibinfo{year}{2012}\natexlab{}.
\newblock \showarticletitle{CityGML--Interoperable semantic 3D city models}.
\newblock \bibinfo{journal}{\emph{ISPRS Journal of Photogrammetry and Remote
  Sensing}}  \bibinfo{volume}{71} (\bibinfo{year}{2012}),
  \bibinfo{pages}{12--33}.
\newblock


\bibitem[Holzmann et~al\mbox{.}(2017)]%
        {holzmann2017plane}
\bibfield{author}{\bibinfo{person}{Thomas Holzmann}, \bibinfo{person}{Martin~R
  Oswald}, \bibinfo{person}{Marc Pollefeys}, \bibinfo{person}{Friedrich
  Fraundorfer}, {and} \bibinfo{person}{Horst Bischof}.}
  \bibinfo{year}{2017}\natexlab{}.
\newblock \showarticletitle{Plane-based surface regularization for urban 3d
  construction}. In \bibinfo{booktitle}{\emph{Proceedings 28th British Machine
  Vision Conference, 2017 (BMVC)}}. \bibinfo{pages}{1--9}.
\newblock


\bibitem[Jain et~al\mbox{.}(2021)]%
        {jain2021deep}
\bibfield{author}{\bibinfo{person}{Arjit Jain}, \bibinfo{person}{Sunita
  Sarawagi}, {and} \bibinfo{person}{Prithviraj Sen}.}
  \bibinfo{year}{2021}\natexlab{}.
\newblock \showarticletitle{Deep indexed active learning for matching
  heterogeneous entity representations}.
\newblock \bibinfo{journal}{\emph{Proceedings of the VLDB Endowment}}
  \bibinfo{volume}{15}, \bibinfo{number}{1} (\bibinfo{year}{2021}),
  \bibinfo{pages}{31--45}.
\newblock


\bibitem[Jaljolie et~al\mbox{.}(2021)]%
        {jaljolie2021topological}
\bibfield{author}{\bibinfo{person}{Ruba Jaljolie}, \bibinfo{person}{Kirsikka
  Riekkinen}, {and} \bibinfo{person}{Sagi Dalyot}.}
  \bibinfo{year}{2021}\natexlab{}.
\newblock \showarticletitle{A topological-based approach for determining
  spatial relationships of complex volumetric parcels in land administration
  systems}.
\newblock \bibinfo{journal}{\emph{Land Use Policy}}  \bibinfo{volume}{109}
  (\bibinfo{year}{2021}), \bibinfo{pages}{105637}.
\newblock


\bibitem[Jaljolie et~al\mbox{.}(2018)]%
        {jaljolie2018spatial}
\bibfield{author}{\bibinfo{person}{Ruba Jaljolie}, \bibinfo{person}{Peter
  Van~Oosterom}, {and} \bibinfo{person}{Sagi Dalyot}.}
  \bibinfo{year}{2018}\natexlab{}.
\newblock \showarticletitle{Spatial data structure and functionalities for 3d
  land management system implementation: Israel case study}.
\newblock \bibinfo{journal}{\emph{Isprs international journal of
  Geo-information}} \bibinfo{volume}{7}, \bibinfo{number}{1}
  (\bibinfo{year}{2018}), \bibinfo{pages}{10}.
\newblock


\bibitem[Jaro(1989)]%
        {jaro1989advances}
\bibfield{author}{\bibinfo{person}{Matthew~A Jaro}.}
  \bibinfo{year}{1989}\natexlab{}.
\newblock \showarticletitle{Advances in record-linkage methodology as applied
  to matching the 1985 census of Tampa, Florida}.
\newblock \bibinfo{journal}{\emph{J. Amer. Statist. Assoc.}}
  \bibinfo{volume}{84}, \bibinfo{number}{406} (\bibinfo{year}{1989}),
  \bibinfo{pages}{414--420}.
\newblock


\bibitem[Jaro(1995)]%
        {jaro1995probabilistic}
\bibfield{author}{\bibinfo{person}{Matthew~A Jaro}.}
  \bibinfo{year}{1995}\natexlab{}.
\newblock \showarticletitle{Probabilistic linkage of large public health data
  files}.
\newblock \bibinfo{journal}{\emph{Statistics in medicine}}
  \bibinfo{volume}{14}, \bibinfo{number}{5-7} (\bibinfo{year}{1995}),
  \bibinfo{pages}{491--498}.
\newblock


\bibitem[Johnson et~al\mbox{.}(2019)]%
        {johnson2019billion}
\bibfield{author}{\bibinfo{person}{Jeff Johnson}, \bibinfo{person}{Matthijs
  Douze}, {and} \bibinfo{person}{Herv{\'e} J{\'e}gou}.}
  \bibinfo{year}{2019}\natexlab{}.
\newblock \showarticletitle{Billion-scale similarity search with GPUs}.
\newblock \bibinfo{journal}{\emph{IEEE Transactions on Big Data}}
  \bibinfo{volume}{7}, \bibinfo{number}{3} (\bibinfo{year}{2019}),
  \bibinfo{pages}{535--547}.
\newblock


\bibitem[Joty and Tang(2018)]%
        {joty2018distributed}
\bibfield{author}{\bibinfo{person}{Muhammad Ebraheem Saravanan
  Thirumuruganathan~Shafiq Joty} {and} \bibinfo{person}{Mourad Ouzzani~Nan
  Tang}.} \bibinfo{year}{2018}\natexlab{}.
\newblock \showarticletitle{Distributed Representations of Tuples for Entity
  Resolution}.
\newblock \bibinfo{journal}{\emph{Proceedings of the VLDB Endowment}}
  \bibinfo{volume}{11}, \bibinfo{number}{11} (\bibinfo{year}{2018}).
\newblock


\bibitem[Jovanovi{\'c} et~al\mbox{.}(2020)]%
        {jovanovic2020building}
\bibfield{author}{\bibinfo{person}{Du{\v{s}}an Jovanovi{\'c}},
  \bibinfo{person}{Stevan Milovanov}, \bibinfo{person}{Igor Ruskovski},
  \bibinfo{person}{Miro Govedarica}, \bibinfo{person}{Dubravka Sladi{\'c}},
  \bibinfo{person}{Aleksandra Radulovi{\'c}}, {and} \bibinfo{person}{Vladimir
  Paji{\'c}}.} \bibinfo{year}{2020}\natexlab{}.
\newblock \showarticletitle{Building virtual 3D city model for smart cities
  applications: A case study on campus area of the university of novi sad}.
\newblock \bibinfo{journal}{\emph{ISPRS International Journal of
  Geo-Information}} \bibinfo{volume}{9}, \bibinfo{number}{8}
  (\bibinfo{year}{2020}), \bibinfo{pages}{476}.
\newblock


\bibitem[Kang et~al\mbox{.}(2007)]%
        {kang2007geoddupe}
\bibfield{author}{\bibinfo{person}{Hyunmo Kang}, \bibinfo{person}{Vivek
  Sehgal}, {and} \bibinfo{person}{Lise Getoor}.}
  \bibinfo{year}{2007}\natexlab{}.
\newblock \showarticletitle{Geoddupe: A novel interface for interactive entity
  resolution in geospatial data}. In \bibinfo{booktitle}{\emph{2007 11th
  International Conference Information Visualization (IV'07)}}. IEEE,
  \bibinfo{pages}{489--496}.
\newblock


\bibitem[Kenig and Gal(2013)]%
        {kenig2013mfiblocks}
\bibfield{author}{\bibinfo{person}{Batya Kenig} {and} \bibinfo{person}{Avigdor
  Gal}.} \bibinfo{year}{2013}\natexlab{}.
\newblock \showarticletitle{MFIBlocks: An effective blocking algorithm for
  entity resolution}.
\newblock \bibinfo{journal}{\emph{Information Systems}} \bibinfo{volume}{38},
  \bibinfo{number}{6} (\bibinfo{year}{2013}), \bibinfo{pages}{908--926}.
\newblock


\bibitem[Khodizadeh-Nahari et~al\mbox{.}(2021)]%
        {khodizadeh2021novel}
\bibfield{author}{\bibinfo{person}{Mohammad Khodizadeh-Nahari},
  \bibinfo{person}{Nasser Ghadiri}, \bibinfo{person}{Ahmad Baraani-Dastjerdi},
  {and} \bibinfo{person}{J{\"o}rg-R{\"u}diger Sack}.}
  \bibinfo{year}{2021}\natexlab{}.
\newblock \showarticletitle{A novel similarity measure for spatial entity
  resolution based on data granularity model: Managing inconsistencies in place
  descriptions}.
\newblock \bibinfo{journal}{\emph{Applied Intelligence}} \bibinfo{volume}{51},
  \bibinfo{number}{8} (\bibinfo{year}{2021}), \bibinfo{pages}{6104--6123}.
\newblock


\bibitem[Konda et~al\mbox{.}(2016)]%
        {konda2016magellan}
\bibfield{author}{\bibinfo{person}{Pradap Konda} {et~al\mbox{.}}}
  \bibinfo{year}{2016}\natexlab{}.
\newblock \showarticletitle{Magellan: Toward building entity matching
  management systems}.
\newblock \bibinfo{journal}{\emph{Proceedings of the VLDB Endowment}}
  \bibinfo{volume}{9}, \bibinfo{number}{12} (\bibinfo{year}{2016}),
  \bibinfo{pages}{1197--1208}.
\newblock


\bibitem[Labetski et~al\mbox{.}(2023)]%
        {labetski20233d}
\bibfield{author}{\bibinfo{person}{Anna Labetski}, \bibinfo{person}{Stelios
  Vitalis}, \bibinfo{person}{Filip Biljecki}, \bibinfo{person}{Ken
  Arroyo~Ohori}, {and} \bibinfo{person}{Jantien Stoter}.}
  \bibinfo{year}{2023}\natexlab{}.
\newblock \showarticletitle{3D building metrics for urban morphology}.
\newblock \bibinfo{journal}{\emph{International Journal of Geographical
  Information Science}} \bibinfo{volume}{37}, \bibinfo{number}{1}
  (\bibinfo{year}{2023}), \bibinfo{pages}{36--67}.
\newblock


\bibitem[Ledoux et~al\mbox{.}(2019)]%
        {ledoux2019cityjson}
\bibfield{author}{\bibinfo{person}{Hugo Ledoux}, \bibinfo{person}{Ken
  Arroyo~Ohori}, \bibinfo{person}{Kavisha Kumar}, \bibinfo{person}{Bal{\'a}zs
  Dukai}, \bibinfo{person}{Anna Labetski}, {and} \bibinfo{person}{Stelios
  Vitalis}.} \bibinfo{year}{2019}\natexlab{}.
\newblock \showarticletitle{CityJSON: A compact and easy-to-use encoding of the
  CityGML data model}.
\newblock \bibinfo{journal}{\emph{Open Geospatial Data, Software and
  Standards}} \bibinfo{volume}{4}, \bibinfo{number}{1} (\bibinfo{year}{2019}),
  \bibinfo{pages}{1--12}.
\newblock


\bibitem[Levenshtein(1966)]%
        {levenshtein1966binary}
\bibfield{author}{\bibinfo{person}{Vladimir~I Levenshtein}.}
  \bibinfo{year}{1966}\natexlab{}.
\newblock \showarticletitle{Binary codes capable of correcting deletions,
  insertions, and reversals}. In \bibinfo{booktitle}{\emph{Soviet physics
  doklady}}, Vol.~\bibinfo{volume}{10}. \bibinfo{pages}{707--710}.
\newblock


\bibitem[Li et~al\mbox{.}(2020b)]%
        {li2020survey}
\bibfield{author}{\bibinfo{person}{Bo-Han Li}, \bibinfo{person}{Yi Liu},
  \bibinfo{person}{An-Man Zhang}, \bibinfo{person}{Wen-Huan Wang}, {and}
  \bibinfo{person}{Shuo Wan}.} \bibinfo{year}{2020}\natexlab{b}.
\newblock \showarticletitle{A survey on blocking technology of entity
  resolution}.
\newblock \bibinfo{journal}{\emph{Journal of Computer Science and Technology}}
  \bibinfo{volume}{35}, \bibinfo{number}{4} (\bibinfo{year}{2020}),
  \bibinfo{pages}{769--793}.
\newblock


\bibitem[Li et~al\mbox{.}(2020a)]%
        {li2020deep}
\bibfield{author}{\bibinfo{person}{Yuliang Li}, \bibinfo{person}{Jinfeng Li},
  \bibinfo{person}{Yoshihiko Suhara}, \bibinfo{person}{AnHai Doan}, {and}
  \bibinfo{person}{Wang-Chiew Tan}.} \bibinfo{year}{2020}\natexlab{a}.
\newblock \showarticletitle{Deep entity matching with pre-trained language
  models}.
\newblock \bibinfo{journal}{\emph{Proceedings of the VLDB Endowment}}
  \bibinfo{volume}{14}, \bibinfo{number}{1} (\bibinfo{year}{2020}),
  \bibinfo{pages}{50--60}.
\newblock


\bibitem[Li et~al\mbox{.}(2021)]%
        {li2021deep}
\bibfield{author}{\bibinfo{person}{Yuliang Li}, \bibinfo{person}{Jinfeng Li},
  \bibinfo{person}{Yoshihiko Suhara}, \bibinfo{person}{Jin Wang},
  \bibinfo{person}{Wataru Hirota}, {and} \bibinfo{person}{Wang-Chiew Tan}.}
  \bibinfo{year}{2021}\natexlab{}.
\newblock \showarticletitle{Deep entity matching: Challenges and
  opportunities}.
\newblock \bibinfo{journal}{\emph{Journal of Data and Information Quality
  (JDIQ)}} \bibinfo{volume}{13}, \bibinfo{number}{1} (\bibinfo{year}{2021}),
  \bibinfo{pages}{1--17}.
\newblock


\bibitem[Lin et~al\mbox{.}(1998)]%
        {lin1998information}
\bibfield{author}{\bibinfo{person}{Dekang Lin} {et~al\mbox{.}}}
  \bibinfo{year}{1998}\natexlab{}.
\newblock \showarticletitle{An information-theoretic definition of similarity}.
  In \bibinfo{booktitle}{\emph{ICML}}, Vol.~\bibinfo{volume}{98}.
  \bibinfo{pages}{296--304}.
\newblock


\bibitem[Manning and Mullahy(2001)]%
        {manning2001estimating}
\bibfield{author}{\bibinfo{person}{Willard~G Manning} {and}
  \bibinfo{person}{John Mullahy}.} \bibinfo{year}{2001}\natexlab{}.
\newblock \showarticletitle{Estimating log models: to transform or not to
  transform?}
\newblock \bibinfo{journal}{\emph{Journal of health economics}}
  \bibinfo{volume}{20}, \bibinfo{number}{4} (\bibinfo{year}{2001}),
  \bibinfo{pages}{461--494}.
\newblock


\bibitem[Manzini et~al\mbox{.}(2024)]%
        {manzini2024crasar}
\bibfield{author}{\bibinfo{person}{Thomas Manzini}, \bibinfo{person}{Priyankari
  Perali}, \bibinfo{person}{Raisa Karnik}, {and} \bibinfo{person}{Robin
  Murphy}.} \bibinfo{year}{2024}\natexlab{}.
\newblock \showarticletitle{Crasar-u-droids: A large scale benchmark dataset
  for building alignment and damage assessment in georectified suas imagery}.
\newblock \bibinfo{journal}{\emph{arXiv preprint arXiv:2407.17673}}
  (\bibinfo{year}{2024}).
\newblock


\bibitem[McElfresh et~al\mbox{.}(2023)]%
        {mcelfresh2023neural}
\bibfield{author}{\bibinfo{person}{Duncan McElfresh}, \bibinfo{person}{Sujay
  Khandagale}, \bibinfo{person}{Jonathan Valverde}, \bibinfo{person}{Vishak
  Prasad~C}, \bibinfo{person}{Ganesh Ramakrishnan}, \bibinfo{person}{Micah
  Goldblum}, {and} \bibinfo{person}{Colin White}.}
  \bibinfo{year}{2023}\natexlab{}.
\newblock \showarticletitle{When do neural nets outperform boosted trees on
  tabular data?}
\newblock \bibinfo{journal}{\emph{Advances in Neural Information Processing
  Systems}}  \bibinfo{volume}{36} (\bibinfo{year}{2023}),
  \bibinfo{pages}{76336--76369}.
\newblock


\bibitem[Morana et~al\mbox{.}({[n.\,d.]})]%
        {morana2014geobench}
\bibfield{author}{\bibinfo{person}{Anthony Morana}, \bibinfo{person}{Thomas
  Morel}, \bibinfo{person}{Bilal Berjawi}, {and} \bibinfo{person}{Fabien
  Duchateau}.} \bibinfo{year}{[n.\,d.]}\natexlab{}.
\newblock \showarticletitle{Geobench: a geospatial integration tool for
  building a spatial entity matching benchmark}. In
  \bibinfo{booktitle}{\emph{Proceedings of the 22nd ACM SIGSPATIAL
  International Conference on Advances in Geographic Information Systems}}.
\newblock


\bibitem[Mudgal et~al\mbox{.}(2018)]%
        {mudgal2018deep}
\bibfield{author}{\bibinfo{person}{Sidharth Mudgal}, \bibinfo{person}{Han Li},
  \bibinfo{person}{Theodoros Rekatsinas}, \bibinfo{person}{AnHai Doan},
  \bibinfo{person}{Youngchoon Park}, \bibinfo{person}{Ganesh Krishnan},
  \bibinfo{person}{Rohit Deep}, \bibinfo{person}{Esteban Arcaute}, {and}
  \bibinfo{person}{Vijay Raghavendra}.} \bibinfo{year}{2018}\natexlab{}.
\newblock \showarticletitle{Deep learning for entity matching: A design space
  exploration}. In \bibinfo{booktitle}{\emph{Proceedings of the 2018
  International Conference on Management of Data}}. \bibinfo{pages}{19--34}.
\newblock


\bibitem[Papadakis et~al\mbox{.}(2015)]%
        {papadakis2015schema}
\bibfield{author}{\bibinfo{person}{George Papadakis}, \bibinfo{person}{George
  Alexiou}, \bibinfo{person}{George Papastefanatos}, {and}
  \bibinfo{person}{Georgia Koutrika}.} \bibinfo{year}{2015}\natexlab{}.
\newblock \showarticletitle{Schema-agnostic vs schema-based configurations for
  blocking methods on homogeneous data}.
\newblock \bibinfo{journal}{\emph{Proceedings of the VLDB Endowment}}
  \bibinfo{volume}{9}, \bibinfo{number}{4} (\bibinfo{year}{2015}),
  \bibinfo{pages}{312--323}.
\newblock


\bibitem[Papadakis et~al\mbox{.}(2011)]%
        {papadakis2011eliminating}
\bibfield{author}{\bibinfo{person}{George Papadakis},
  \bibinfo{person}{Ekaterini Ioannou}, \bibinfo{person}{Claudia Nieder{\'e}e},
  \bibinfo{person}{Themis Palpanas}, {and} \bibinfo{person}{Wolfgang Nejdl}.}
  \bibinfo{year}{2011}\natexlab{}.
\newblock \showarticletitle{Eliminating the redundancy in blocking-based entity
  resolution methods}. In \bibinfo{booktitle}{\emph{Proceedings of the 11th
  annual international ACM/IEEE joint conference on Digital libraries}}.
  \bibinfo{pages}{85--94}.
\newblock


\bibitem[Papadakis et~al\mbox{.}(2019)]%
        {papadakis2019survey}
\bibfield{author}{\bibinfo{person}{George Papadakis},
  \bibinfo{person}{Dimitrios Skoutas}, \bibinfo{person}{Emmanouil Thanos},
  {and} \bibinfo{person}{Themis Palpanas}.} \bibinfo{year}{2019}\natexlab{}.
\newblock \showarticletitle{A survey of blocking and filtering techniques for
  entity resolution}.
\newblock \bibinfo{journal}{\emph{arXiv preprint arXiv:1905.06167}}
  (\bibinfo{year}{2019}).
\newblock


\bibitem[Papadakis et~al\mbox{.}(2020)]%
        {papadakis2020blocking}
\bibfield{author}{\bibinfo{person}{George Papadakis},
  \bibinfo{person}{Dimitrios Skoutas}, \bibinfo{person}{Emmanouil Thanos},
  {and} \bibinfo{person}{Themis Palpanas}.} \bibinfo{year}{2020}\natexlab{}.
\newblock \showarticletitle{Blocking and filtering techniques for entity
  resolution: A survey}.
\newblock \bibinfo{journal}{\emph{ACM Computing Surveys (CSUR)}}
  \bibinfo{volume}{53}, \bibinfo{number}{2} (\bibinfo{year}{2020}),
  \bibinfo{pages}{1--42}.
\newblock


\bibitem[Paulsen et~al\mbox{.}(2023)]%
        {paulsen2023sparkly}
\bibfield{author}{\bibinfo{person}{Derek Paulsen}, \bibinfo{person}{Yash
  Govind}, {and} \bibinfo{person}{AnHai Doan}.}
  \bibinfo{year}{2023}\natexlab{}.
\newblock \showarticletitle{Sparkly: A simple yet surprisingly strong TF/IDF
  blocker for entity matching}.
\newblock \bibinfo{journal}{\emph{Proceedings of the VLDB Endowment}}
  \bibinfo{volume}{16}, \bibinfo{number}{6} (\bibinfo{year}{2023}),
  \bibinfo{pages}{1507--1519}.
\newblock


\bibitem[Pedregosa et~al\mbox{.}(2011)]%
        {pedregosa2011scikit}
\bibfield{author}{\bibinfo{person}{Fabian Pedregosa}, \bibinfo{person}{Ga{\"e}l
  Varoquaux}, \bibinfo{person}{Alexandre Gramfort}, \bibinfo{person}{Vincent
  Michel}, \bibinfo{person}{Bertrand Thirion}, \bibinfo{person}{Olivier
  Grisel}, \bibinfo{person}{Mathieu Blondel}, \bibinfo{person}{Peter
  Prettenhofer}, \bibinfo{person}{Ron Weiss}, \bibinfo{person}{Vincent
  Dubourg}, {et~al\mbox{.}}} \bibinfo{year}{2011}\natexlab{}.
\newblock \showarticletitle{Scikit-learn: Machine learning in Python}.
\newblock \bibinfo{journal}{\emph{the Journal of machine Learning research}}
  \bibinfo{volume}{12} (\bibinfo{year}{2011}), \bibinfo{pages}{2825--2830}.
\newblock


\bibitem[Peeters and Bizer(2023)]%
        {peeters2023using}
\bibfield{author}{\bibinfo{person}{Ralph Peeters} {and}
  \bibinfo{person}{Christian Bizer}.} \bibinfo{year}{2023}\natexlab{}.
\newblock \showarticletitle{Using chatgpt for entity matching}. In
  \bibinfo{booktitle}{\emph{European Conference on Advances in Databases and
  Information Systems}}. Springer, \bibinfo{pages}{221--230}.
\newblock


\bibitem[Peeters et~al\mbox{.}(2023)]%
        {peeters2023entity}
\bibfield{author}{\bibinfo{person}{Ralph Peeters}, \bibinfo{person}{Aaron
  Steiner}, {and} \bibinfo{person}{Christian Bizer}.}
  \bibinfo{year}{2023}\natexlab{}.
\newblock \showarticletitle{Entity matching using large language models}.
\newblock \bibinfo{journal}{\emph{arXiv preprint arXiv:2310.11244}}
  (\bibinfo{year}{2023}).
\newblock


\bibitem[Peeters et~al\mbox{.}(2025)]%
        {peeters2025entity}
\bibfield{author}{\bibinfo{person}{Ralph Peeters}, \bibinfo{person}{Aaron
  Steiner}, {and} \bibinfo{person}{Christian Bizer}.}
  \bibinfo{year}{2025}\natexlab{}.
\newblock \showarticletitle{Entity matching using large language models}.
\newblock \bibinfo{journal}{\emph{OpenProceedings}}  \bibinfo{volume}{2}
  (\bibinfo{year}{2025}), \bibinfo{pages}{529--541}.
\newblock


\bibitem[Peters et~al\mbox{.}(2022)]%
        {peters2022automated}
\bibfield{author}{\bibinfo{person}{Ravi Peters}, \bibinfo{person}{Bal{\'a}zs
  Dukai}, \bibinfo{person}{Stelios Vitalis}, \bibinfo{person}{Jordi van
  Liempt}, {and} \bibinfo{person}{Jantien Stoter}.}
  \bibinfo{year}{2022}\natexlab{}.
\newblock \showarticletitle{Automated 3D reconstruction of LoD2 and LoD1 models
  for all 10 million buildings of the Netherlands}.
\newblock \bibinfo{journal}{\emph{Photogrammetric Engineering \& Remote
  Sensing}} \bibinfo{volume}{88}, \bibinfo{number}{3} (\bibinfo{year}{2022}),
  \bibinfo{pages}{165--170}.
\newblock


\bibitem[Radford et~al\mbox{.}(2021)]%
        {radford2021learning}
\bibfield{author}{\bibinfo{person}{Alec Radford}, \bibinfo{person}{Jong~Wook
  Kim}, \bibinfo{person}{Chris Hallacy}, \bibinfo{person}{Aditya Ramesh},
  \bibinfo{person}{Gabriel Goh}, \bibinfo{person}{Sandhini Agarwal},
  \bibinfo{person}{Girish Sastry}, \bibinfo{person}{Amanda Askell},
  \bibinfo{person}{Pamela Mishkin}, \bibinfo{person}{Jack Clark},
  {et~al\mbox{.}}} \bibinfo{year}{2021}\natexlab{}.
\newblock \showarticletitle{Learning transferable visual models from natural
  language supervision}. In \bibinfo{booktitle}{\emph{International conference
  on machine learning}}. PmLR, \bibinfo{pages}{8748--8763}.
\newblock


\bibitem[Ram and Sinha(2019)]%
        {ram2019revisiting}
\bibfield{author}{\bibinfo{person}{Parikshit Ram} {and}
  \bibinfo{person}{Kaushik Sinha}.} \bibinfo{year}{2019}\natexlab{}.
\newblock \showarticletitle{Revisiting kd-tree for nearest neighbor search}. In
  \bibinfo{booktitle}{\emph{Proceedings of the 25th acm sigkdd international
  conference on knowledge discovery \& data mining}}.
  \bibinfo{pages}{1378--1388}.
\newblock


\bibitem[Sehgal et~al\mbox{.}(2006)]%
        {sehgal2006entity}
\bibfield{author}{\bibinfo{person}{Vivek Sehgal}, \bibinfo{person}{Lise
  Getoor}, {and} \bibinfo{person}{Peter~D Viechnicki}.}
  \bibinfo{year}{2006}\natexlab{}.
\newblock \showarticletitle{Entity resolution in geospatial data integration}.
  In \bibinfo{booktitle}{\emph{Proceedings of the 14th annual ACM international
  symposium on Advances in geographic information systems}}.
  \bibinfo{pages}{83--90}.
\newblock


\bibitem[Shah et~al\mbox{.}(2021)]%
        {shah2021gem}
\bibfield{author}{\bibinfo{person}{Setu Shah}, \bibinfo{person}{Vamsi Meduri},
  {and} \bibinfo{person}{Mohamed Sarwat}.} \bibinfo{year}{2021}\natexlab{}.
\newblock \showarticletitle{GEM: An efficient entity matching framework for
  geospatial data}. In \bibinfo{booktitle}{\emph{Proceedings of the 29th
  International Conference on Advances in Geographic Information Systems}}.
  \bibinfo{pages}{346--349}.
\newblock


\bibitem[Singh et~al\mbox{.}(2017)]%
        {singh2017synthesizing}
\bibfield{author}{\bibinfo{person}{Rohit Singh},
  \bibinfo{person}{Venkata~Vamsikrishna Meduri}, \bibinfo{person}{Ahmed
  Elmagarmid}, \bibinfo{person}{Samuel Madden}, \bibinfo{person}{Paolo
  Papotti}, \bibinfo{person}{Jorge-Arnulfo Quian{\'e}-Ruiz},
  \bibinfo{person}{Armando Solar-Lezama}, {and} \bibinfo{person}{Nan Tang}.}
  \bibinfo{year}{2017}\natexlab{}.
\newblock \showarticletitle{Synthesizing entity matching rules by examples}.
\newblock \bibinfo{journal}{\emph{Proceedings of the VLDB Endowment}}
  \bibinfo{volume}{11}, \bibinfo{number}{2} (\bibinfo{year}{2017}),
  \bibinfo{pages}{189--202}.
\newblock


\bibitem[Singh et~al\mbox{.}(2013)]%
        {singh2013virtual}
\bibfield{author}{\bibinfo{person}{Surendra~Pal Singh}, \bibinfo{person}{Kamal
  Jain}, {and} \bibinfo{person}{V~Ravibabu Mandla}.}
  \bibinfo{year}{2013}\natexlab{}.
\newblock \showarticletitle{Virtual 3D city modeling: techniques and
  applications}.
\newblock \bibinfo{journal}{\emph{The International Archives of the
  Photogrammetry, Remote Sensing and Spatial Information Sciences}}
  \bibinfo{volume}{40} (\bibinfo{year}{2013}), \bibinfo{pages}{73--91}.
\newblock


\bibitem[Singla and Domingos(2006)]%
        {singla2006entity}
\bibfield{author}{\bibinfo{person}{Parag Singla} {and} \bibinfo{person}{Pedro
  Domingos}.} \bibinfo{year}{2006}\natexlab{}.
\newblock \showarticletitle{Entity resolution with markov logic}. In
  \bibinfo{booktitle}{\emph{Sixth International Conference on Data Mining
  (ICDM'06)}}. IEEE, \bibinfo{pages}{572--582}.
\newblock


\bibitem[Tabarro et~al\mbox{.}(2017)]%
        {tabarro2017webgis}
\bibfield{author}{\bibinfo{person}{PG Tabarro}, \bibinfo{person}{Jacynthe
  Pouliot}, \bibinfo{person}{Richard Fortier}, {and} \bibinfo{person}{L-M
  Losier}.} \bibinfo{year}{2017}\natexlab{}.
\newblock \showarticletitle{A WebGIS to support GPR 3D data acquisition: A
  first step for the integration of underground utility networks in 3D city
  models}.
\newblock \bibinfo{journal}{\emph{The International Archives of the
  Photogrammetry, Remote Sensing and Spatial Information Sciences}}
  \bibinfo{volume}{42} (\bibinfo{year}{2017}), \bibinfo{pages}{43--48}.
\newblock


\bibitem[Tam et~al\mbox{.}(2012)]%
        {tam2012registration}
\bibfield{author}{\bibinfo{person}{Gary~KL Tam}, \bibinfo{person}{Zhi-Quan
  Cheng}, \bibinfo{person}{Yu-Kun Lai}, \bibinfo{person}{Frank~C Langbein},
  \bibinfo{person}{Yonghuai Liu}, \bibinfo{person}{David Marshall},
  \bibinfo{person}{Ralph~R Martin}, \bibinfo{person}{Xian-Fang Sun}, {and}
  \bibinfo{person}{Paul~L Rosin}.} \bibinfo{year}{2012}\natexlab{}.
\newblock \showarticletitle{Registration of 3D point clouds and meshes: A
  survey from rigid to nonrigid}.
\newblock \bibinfo{journal}{\emph{IEEE transactions on visualization and
  computer graphics}} \bibinfo{volume}{19}, \bibinfo{number}{7}
  (\bibinfo{year}{2012}), \bibinfo{pages}{1199--1217}.
\newblock


\bibitem[Thirumuruganathan et~al\mbox{.}(2021)]%
        {thirumuruganathan2021deep}
\bibfield{author}{\bibinfo{person}{Saravanan Thirumuruganathan},
  \bibinfo{person}{Han Li}, \bibinfo{person}{Nan Tang}, \bibinfo{person}{Mourad
  Ouzzani}, \bibinfo{person}{Yash Govind}, \bibinfo{person}{Derek Paulsen},
  \bibinfo{person}{Glenn Fung}, {and} \bibinfo{person}{AnHai Doan}.}
  \bibinfo{year}{2021}\natexlab{}.
\newblock \showarticletitle{Deep learning for blocking in entity matching: a
  design space exploration}.
\newblock \bibinfo{journal}{\emph{Proceedings of the VLDB Endowment}}
  \bibinfo{volume}{14}, \bibinfo{number}{11} (\bibinfo{year}{2021}),
  \bibinfo{pages}{2459--2472}.
\newblock


\bibitem[Ulusoy and Mundy(2014)]%
        {ulusoy2014image}
\bibfield{author}{\bibinfo{person}{Ali~Osman Ulusoy} {and}
  \bibinfo{person}{Joseph~L Mundy}.} \bibinfo{year}{2014}\natexlab{}.
\newblock \showarticletitle{Image-based 4-d reconstruction using 3-d change
  detection}. In \bibinfo{booktitle}{\emph{European Conference on Computer
  Vision}}. Springer, \bibinfo{pages}{31--45}.
\newblock


\bibitem[Verykokou et~al\mbox{.}(2016)]%
        {verykokou2016uav}
\bibfield{author}{\bibinfo{person}{Styliani Verykokou},
  \bibinfo{person}{Anastasios Doulamis}, \bibinfo{person}{George Athanasiou},
  \bibinfo{person}{Charalabos Ioannidis}, {and} \bibinfo{person}{Angelos
  Amditis}.} \bibinfo{year}{2016}\natexlab{}.
\newblock \showarticletitle{UAV-based 3D modelling of disaster scenes for Urban
  Search and Rescue}. In \bibinfo{booktitle}{\emph{2016 IEEE International
  Conference on Imaging Systems and Techniques (IST)}}. IEEE,
  \bibinfo{pages}{106--111}.
\newblock


\bibitem[Wang et~al\mbox{.}(2022)]%
        {wang2022promptem}
\bibfield{author}{\bibinfo{person}{Pengfei Wang}, \bibinfo{person}{Xiaocan
  Zeng}, \bibinfo{person}{Lu Chen}, \bibinfo{person}{Fan Ye},
  \bibinfo{person}{Yuren Mao}, \bibinfo{person}{Junhao Zhu}, {and}
  \bibinfo{person}{Yunjun Gao}.} \bibinfo{year}{2022}\natexlab{}.
\newblock \showarticletitle{Promptem: prompt-tuning for low-resource
  generalized entity matching}.
\newblock \bibinfo{journal}{\emph{arXiv preprint arXiv:2207.04802}}
  (\bibinfo{year}{2022}).
\newblock


\bibitem[Winkler(2002)]%
        {winkler2002methods}
\bibfield{author}{\bibinfo{person}{William~E Winkler}.}
  \bibinfo{year}{2002}\natexlab{}.
\newblock \bibinfo{booktitle}{\emph{Methods for record linkage and bayesian
  networks}}.
\newblock \bibinfo{type}{{T}echnical {R}eport}.
  \bibinfo{institution}{Statistical Research Division, US Census Bureau}.
\newblock


\bibitem[Wu et~al\mbox{.}(2020)]%
        {wu2020zeroer}
\bibfield{author}{\bibinfo{person}{Renzhi Wu}, \bibinfo{person}{Sanya Chaba},
  \bibinfo{person}{Saurabh Sawlani}, \bibinfo{person}{Xu Chu}, {and}
  \bibinfo{person}{Saravanan Thirumuruganathan}.}
  \bibinfo{year}{2020}\natexlab{}.
\newblock \showarticletitle{Zeroer: Entity resolution using zero labeled
  examples}. In \bibinfo{booktitle}{\emph{Proceedings of the 2020 ACM SIGMOD
  International Conference on Management of Data}}.
  \bibinfo{pages}{1149--1164}.
\newblock


\bibitem[Xu et~al\mbox{.}(2021)]%
        {xu2021volumetric}
\bibfield{author}{\bibinfo{person}{Ningli Xu}, \bibinfo{person}{Debao Huang},
  \bibinfo{person}{Shuang Song}, \bibinfo{person}{Xiao Ling},
  \bibinfo{person}{Chris Strasbaugh}, \bibinfo{person}{Alper Yilmaz},
  \bibinfo{person}{Halil Sezen}, {and} \bibinfo{person}{Rongjun Qin}.}
  \bibinfo{year}{2021}\natexlab{}.
\newblock \showarticletitle{A volumetric change detection framework using UAV
  oblique photogrammetry--a case study of ultra-high-resolution monitoring of
  progressive building collapse}.
\newblock \bibinfo{journal}{\emph{International Journal of Digital Earth}}
  \bibinfo{volume}{14}, \bibinfo{number}{11} (\bibinfo{year}{2021}),
  \bibinfo{pages}{1705--1720}.
\newblock


\bibitem[Yu et~al\mbox{.}(2018)]%
        {yu2018big}
\bibfield{author}{\bibinfo{person}{Manzhu Yu}, \bibinfo{person}{Chaowei Yang},
  {and} \bibinfo{person}{Yun Li}.} \bibinfo{year}{2018}\natexlab{}.
\newblock \showarticletitle{Big data in natural disaster management: a review}.
\newblock \bibinfo{journal}{\emph{Geosciences}} \bibinfo{volume}{8},
  \bibinfo{number}{5} (\bibinfo{year}{2018}), \bibinfo{pages}{165}.
\newblock


\bibitem[Zhang et~al\mbox{.}(2020)]%
        {zhang2020autoblock}
\bibfield{author}{\bibinfo{person}{Wei Zhang}, \bibinfo{person}{Hao Wei},
  \bibinfo{person}{Bunyamin Sisman}, \bibinfo{person}{Xin~Luna Dong},
  \bibinfo{person}{Christos Faloutsos}, {and} \bibinfo{person}{Davd Page}.}
  \bibinfo{year}{2020}\natexlab{}.
\newblock \showarticletitle{Autoblock: A hands-off blocking framework for
  entity matching}. In \bibinfo{booktitle}{\emph{Proceedings of the 13th
  International Conference on Web Search and Data Mining}}.
  \bibinfo{pages}{744--752}.
\newblock


\bibitem[Zhuo et~al\mbox{.}(2017)]%
        {zhuo2017automatic}
\bibfield{author}{\bibinfo{person}{Xiangyu Zhuo}, \bibinfo{person}{Tobias
  Koch}, \bibinfo{person}{Franz Kurz}, \bibinfo{person}{Friedrich Fraundorfer},
  {and} \bibinfo{person}{Peter Reinartz}.} \bibinfo{year}{2017}\natexlab{}.
\newblock \showarticletitle{Automatic UAV image geo-registration by matching
  UAV images to georeferenced image data}.
\newblock \bibinfo{journal}{\emph{Remote Sensing}} \bibinfo{volume}{9},
  \bibinfo{number}{4} (\bibinfo{year}{2017}), \bibinfo{pages}{376}.
\newblock


\end{thebibliography}
